\documentclass[sn-mathphys-num]{sn-jnl}

\usepackage{graphicx}%
\usepackage{multirow}%
\usepackage{amsmath,amssymb,amsfonts}%
\usepackage{amsthm}%
\usepackage{mathrsfs}%
\usepackage[title]{appendix}%
\usepackage{xcolor}%
\usepackage{textcomp}%
\usepackage{manyfoot}%
\usepackage{booktabs}%
\usepackage{algorithm}%
\usepackage{algorithmicx}%
\usepackage{algpseudocode}%
\usepackage{listings}%
\usepackage{longtable}
\usepackage{tabularx}
\usepackage{array}
\usepackage{booktabs}
\usepackage{rotating}
\usepackage{makecell}
\usepackage{amssymb} 
\usepackage{soul}
\usepackage{lineno}
\usepackage{caption}
\newcolumntype{L}[1]{>{\raggedright\arraybackslash}p{#1}}
\newcolumntype{C}[1]{>{\centering\arraybackslash}p{#1}}
\newcolumntype{R}[1]{>{\raggedleft\arraybackslash}p{#1}}
\newcolumntype{Y}{>{\raggedright\arraybackslash}X} 
\theoremstyle{thmstyleone}%
\theoremstyle{thmstyletwo}%

\theoremstyle{thmstylethree}%

\begin{document}

\title[Article Title]{Vision Transformers in Domain Adaptation and Domain Generalization: A study of Robustness}


\author[1]{\fnm{Shadi} \sur{Alijani}}\email{shadialijani@uvic.ca}

\author[1]{\fnm{Jamil} \sur{Fayyad}}\email{jfayyad@uvic.ca}

\author*[1]{\fnm{Homayoun} \sur{Najjaran}}\email{najjaran@uvic.ca}

\affil*[1]{\orgname{University of Victoria}, \orgaddress{\street{800 Finnerty Road}, \city{Victoria}, \postcode{V8P 5C2}, \state{BC}, \country{Canada}}}




\abstract{Deep learning models are often evaluated in scenarios where the data distribution is different from those used in the training and validation phases. The discrepancy presents a challenge for accurately predicting the performance of models once deployed on the target distribution. Domain adaptation and generalization are widely recognized as effective strategies for addressing such shifts, thereby ensuring reliable performance. The recent promising results in applying vision transformers in computer vision tasks, coupled with advancements in self-attention mechanisms, have demonstrated their significant potential for robustness and generalization in handling distribution shifts. Motivated by the increased interest from the research community, our paper investigates the deployment of vision transformers in domain adaptation and domain generalization scenarios. For domain adaptation methods, we categorize research into feature-level, instance-level, model-level adaptations, and hybrid approaches, along with other categorizations with respect to diverse strategies for enhancing domain adaptation. Similarly, for domain generalization, we categorize research into multi-domain learning, meta-learning, regularization techniques, and data augmentation strategies. We further classify diverse strategies in research, underscoring the various approaches researchers have taken to address distribution shifts by integrating vision transformers. The inclusion of comprehensive tables summarizing these categories is a distinct feature of our work, offering valuable insights for researchers. These findings highlight the versatility of vision transformers in managing distribution shifts, crucial for real-world applications, especially in critical safety and decision-making scenarios.}




\maketitle
\section{Introduction}\label{sec1}

Convolutional Neural Networks (CNNs) are a cornerstone of computer vision algorithms, largely owing to their proficiency in managing spatial relationships and maintaining invariance to input translations. Their widespread success in object recognition tasks can be attributed to advantageous inductive biases, such as translation equivalence, which enable them to effectively identify and process visual patterns. The foundational concept of using convolutions in neural networks was initiated by Fukushima's development of the Neocognitron \cite{fukushima1980neocognitron}, a model that introduced the idea of a shift-invariant architecture. This idea was further advanced by LeCun et al., who applied gradient-based learning to document recognition, showcasing the practical applicability of CNNs \cite{lecun1998gradient}. The significant breakthrough in CNNs came with Krizhevsky et al., whose work on ImageNet classification popularized deep convolutional networks \cite{krizhevsky2012imagenet}. Following this, developments such as Szegedy et al.'s deeper convolutional networks \cite{szegedy2015going}, He et al.'s introduction of residual networks \cite{he2016deep}, and Huang et al.'s densely connected networks \cite{huang2017densely} have each contributed unique architectural improvements that enhance model robustness and accuracy. Recent studies like those by Hsieh et al. \cite{hsieh2019robustness} and Tan and Le \cite{tan2019efficientnet} continue to explore the limits of CNN efficiency and robustness, further solidifying the central role of convolutional layers in modern vision networks. These convolutional layers have been further improved with innovations such as residual connections \cite{he2016deep}. Extensive use has led to detailed empirical \cite{szegedy2013intriguing} and analytical evaluations of convolutional networks \cite{girshick2015deformable, battaglia2018relational}. 

Recent advancements, however, have shown the potential of transformers regarding to their self-attention mechanisms that find the global features of the data which provide a more holistic view of the data \cite{schaerf2023art}, reduce inductive bias, and exhibit a high degree of scalability and flexibility. These factors collectively enhance the model's ability to generalize better during testing. 
After their tremendous success in Natural Language Processing (NLP) tasks, transformers are now being actively integrated into computer vision tasks. Pioneering works like Vaswani et al. \cite{vaswani2017attention} introduced transformers, showcasing their efficiency in handling long-range dependencies in data. This approach was extended to language understanding by Devlin et al. \cite{devlin2018bert} with BERT, which dramatically improved the performance of NLP tasks. Brown et al. \cite{brown2020language} further demonstrated the capability of transformers in NLP with their few-shot learning approaches. In the realm of computer vision, Chen et al. \cite{chen2018encoder} and Dosovitskiy et al. \cite{dosovitskiy2020image} adapted transformer architectures to manage spatial hierarchies in images, leading to significant advancements in image segmentation and recognition tasks. Touvron et al. \cite{touvron2021training} explored training data-efficient image transformers, which optimize the transformer architecture for better performance with limited data. Khan et al. \cite{khan2022transformers} provided a comprehensive survey on the application of transformers in vision, encapsulating various models and methodologies that have evolved over time. Adaptformer by Chen et al. \cite{chen2022adaptformer} adapts ViTs for scalable visual recognition, enhancing their adaptability and efficiency across different scales. In terms of integrating vision and language tasks, works like VideoBERT by Sun et al. \cite{sun2019videobert} and ViLBERT by Lu et al. \cite{lu2019vilbert} have been foundational, developing joint models that learn correlated features between video and text. LXMERT by Tan et al. \cite{tan2019lxmert} and UNITER by Chen et al. \cite{chen2019uniter} further refine these approaches, improving the cross-modal understanding necessary for complex tasks involving both vision and language. Finally, Radford et al. \cite{radford2021learning} explored the use of transformers to develop visual models that can be trained using only natural language descriptions, rather than traditional image labels. This approach leverages the rich contextual information available in language to enhance the model's ability to understand and generalize across different visual and textual modalities. By doing so, they are advancing the capacity of models to perform tasks in more diverse and complex environments, effectively bridging the gap between vision and language.

Vision Transformers (ViT) \cite{dosovitskiy2020image}, stands out as a key development in this area, applying a self-attention-based mechanism to sequences of image patches. It achieves competitive performance on the challenging ImageNet classification task \cite{russakovsky2015imagenet}, compared to CNNs. Researchers discovered that existing CNN architectures exhibit limited generalization capabilities when confronted with distribution shift scenarios \cite{hendrycks2021many, bai2021transformers}. Subsequent research, as seen in works like \cite{liu2021swin, wang2021pyramid}, has further expanded the capabilities of transformers, demonstrating impressive performance across various visual benchmarks. This includes applications in different benchmarks including COCO (Common Objects in Context) dataset in object detection and instance segmentation \cite{lin2014microsoft}, as well as ADE20K dataset for semantic segmentation \cite{zhou2017scene}.

As ViTs gain popularity, it becomes crucial to examine the characteristics of the representations they learn \cite{naseer2021intriguing}. This is important in areas like autonomous driving \cite{feng2020deep, fayyad2020deep}, robotics \cite{dhillon2002robot}, and healthcare \cite{ranschaert2019artificial, hemalakshmi2024automated}, where the trustworthy and reliability of these systems are crucial. Recent studies delve into evaluating ViTs’ robustness, focusing not just on standard metrics like accuracy and computational cost, but also on their intrinsic impact on model robustness and generalization, especially in handling Distribution Shifts. In conventional training and testing scenarios, it is assumed that data are independent and identically distributed (IID). However, this assumption often doesn't reflect real-world scenarios. Therefore, exploring the potential of ViTs as the modern vision networks, to adapt to target domains, and generalize and perform well on unseen data, becomes a crucial aspect of machine learning models \cite{zhang2022delving}. 

Recognizing the unique capabilities of ViTs in modern vision tasks highlights the need to assess their performance across varied conditions. In traditional deep learning training and testing scenarios, there is a common assumption that the data are independent and identically distributed (IID). However, any shift in data distribution or domain after training can reduce testing performance \cite{patel2015visual,fayyad2023out, fayyad2024exploiting}. Such IID assumptions often fall short in real-world scenarios, where distribution shifts are prevalent. Thus, the ability of deep learning models to generalize and retain performance across different test domains is crucial for determining their effectiveness \cite{zhang2022delving, angarano2022back}. Exploring the adaptability of ViTs necessitates revisiting \textit{Domain Adaptation (DA)} and \textit{Domain Generalization (DG)}, which are fundamental strategies in machine learning aimed at addressing the challenges posed by distribution shifts between training and testing data, especially when these shifts are pronounced \cite{schwonberg2023augmentation}. Although DA and DG have been traditionally used to overcome such challenges, applying these strategies within the advanced framework of ViTs offers a new approach to examine how these innovative models manage and excel distribution shifts. DA, which provides access to the target domain, and DG, where the target domain remains unseen, represent two approaches to this issue. DA seeks to minimize the discrepancy between specific source and target domains, while DG strives to create a model that remains effective across various unseen domains by utilizing the diversity of multiple source domains during training. This often includes the development of domain-agnostic features that are effective across various domains \cite{wang2022generalizing, wilson2020survey}.

\begin{figure*}
\centering
\includegraphics[width=1\linewidth]{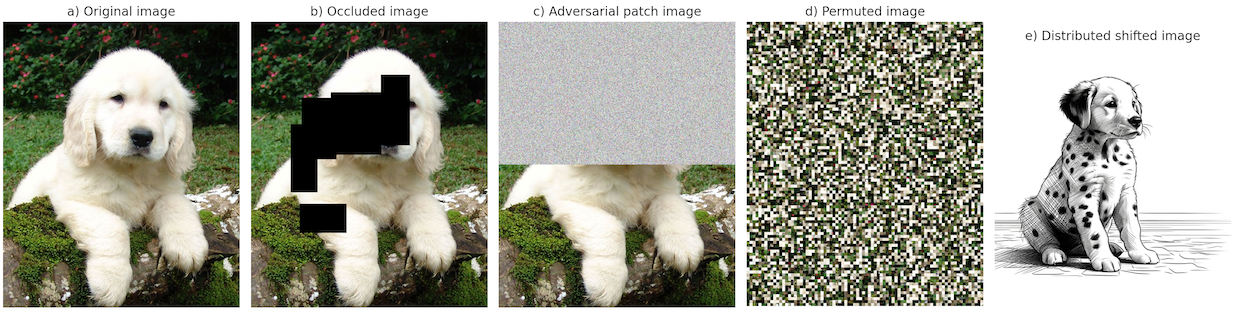}
\caption{Various factors that affect the robustness of deep learning models include: (a) displaying the original image, followed by (b) severe occlusions, (c) adversarial perturbations, (d) patch permutations, and (e) distributional shifts, such as stylization to remove texture cues.}
\label{ViTs vs CNNs}
\end{figure*}

Building on this foundation, research efforts have been directed at enhancing ViTs' generalization capabilities through various methodologies, delving into both DA and DG strategies, and integrating ViTs into the broader deep learning framework. Comparative analyses of ViTs and high-performing CNNs reveal distinct advantages attributable to ViTs. A key differentiation is the dynamic nature of weight computation in ViTs through the self-attention mechanism, contrasting with the static weights learned by CNNs during training. This attribute provides ViTs with a more flexible and adaptable ability \cite{hoyer2022daformer}. ViTs employ multi-head self-attention to intricately parse and interpret contextual information within images, thereby excelling in scenarios involving occlusions, domain variations, and perturbations. They demonstrate remarkable robustness, effectively maintaining accuracy despite image modifications \cite{kim2023improved}. Furthermore, ViTs exhibit a reduced texture bias compared to CNNs, favoring shape recognition which aligns more closely with human visual processing. This proficiency in discerning overall shapes facilitates accurate image categorization without relying on detailed, pixel-level analysis. ViTs' ability to merge various features for image classification enhances their performance across diverse datasets, proving advantageous in both conventional and few-shot learning settings where the model is trained with only a few examples \cite{naseer2021intriguing, gidaris2018unsupervised, raghu2021vision}. Refer to Figure \ref{ViTs vs CNNs} for the challenges prevalent in images. These challenges include severe occlusions, adversarial perturbations, patch permutations, and domain shifts effectively addressed through the flexibility and dynamic receptive fields of self-attention mechanisms \cite{naseer2021intriguing}. Unlike CNNs, which primarily focus on texture \cite{geirhos2018imagenet}, ViTs concentrate more on object shape, enhancing their resistance to texture shifts and benefiting shape recognition, and they are adept at propagating spatial information, an advantage for tasks such as detection and segmentation \cite{naseer2021intriguing, raghu2021vision}.\\

In our comprehensive review, the first of its kind to explore the potential of ViTs in DA and DG scenarios, we have meticulously examined how ViTs adapt to distribution shifts. This study delves into the fundamentals, architecture, and key components of ViTs, offering a unique categorization and analysis of their role in both the theoretical aspects and practical implementations of DA and DG. In this review paper, we explored all the existing papers in this field and developed our own categorizations for the research. Within the context of DA, we categorized the research into feature-level, instance-level, model-level, and hybrid approaches. For DG, our categorization includes multi-domain learning, meta-learning approaches, regularization techniques, and data augmentation strategies.
After the first categorization, we found that there are many studies using ViTs in different categories for DA and DG. The research is somewhat sparse, with researchers applying these modern vision networks in various DA and DG strategies, making it challenging to categorize this diverse research. To address this, we introduced another categorization, providing tables including each study that use different methods. We defined the methods they are using and then, for each study, specified these methods. This dual categorization approach is particularly useful because most of the research employs hybrid methods.
A significant portion of our review is dedicated to showcasing the applications of ViTs beyond image recognition, such as semantic segmentation, action recognition, face analysis, medical imaging, and other emerging fields. This broad spectrum of applications highlights the versatility and potential of ViTs across the vast landscape of computer vision. In our discussion section, we delve into the initial development challenges associated with ViTs, aiming to equip researchers with insightful findings that could steer the direction of future investigations in this area. Furthermore, we outline prospective research paths.

As the field of computer vision advances, especially with the introduction of ViTs, this survey emerges as a promising source for researchers and practitioners alike. We aspire that our findings will stimulate further investigation and innovation in leveraging ViTs for Domain Adaptation and Generalization, thereby overcoming current hurdles and paving new paths in this ever-evolving research domain.

The structure of this paper is as follows: Section \ref{sec:2} introduces the fundamentals and architecture of ViTs. Section \ref{sec:3} assesses the capacity of ViTs to manage distribution shifts, including DA and DG. Section \ref{sec:4} explores various applications of ViTs in computer vision beyond image recognition, particularly their adaptability to distribution shifts. Section \ref{sec:5} wraps up with a comprehensive discussion and conclusion, also suggesting future research directions.

\section{Vision Transformers: Fundamentals and Architecture} \label{sec:2}
\begin{figure*}
\centering
\includegraphics[width=1\linewidth]{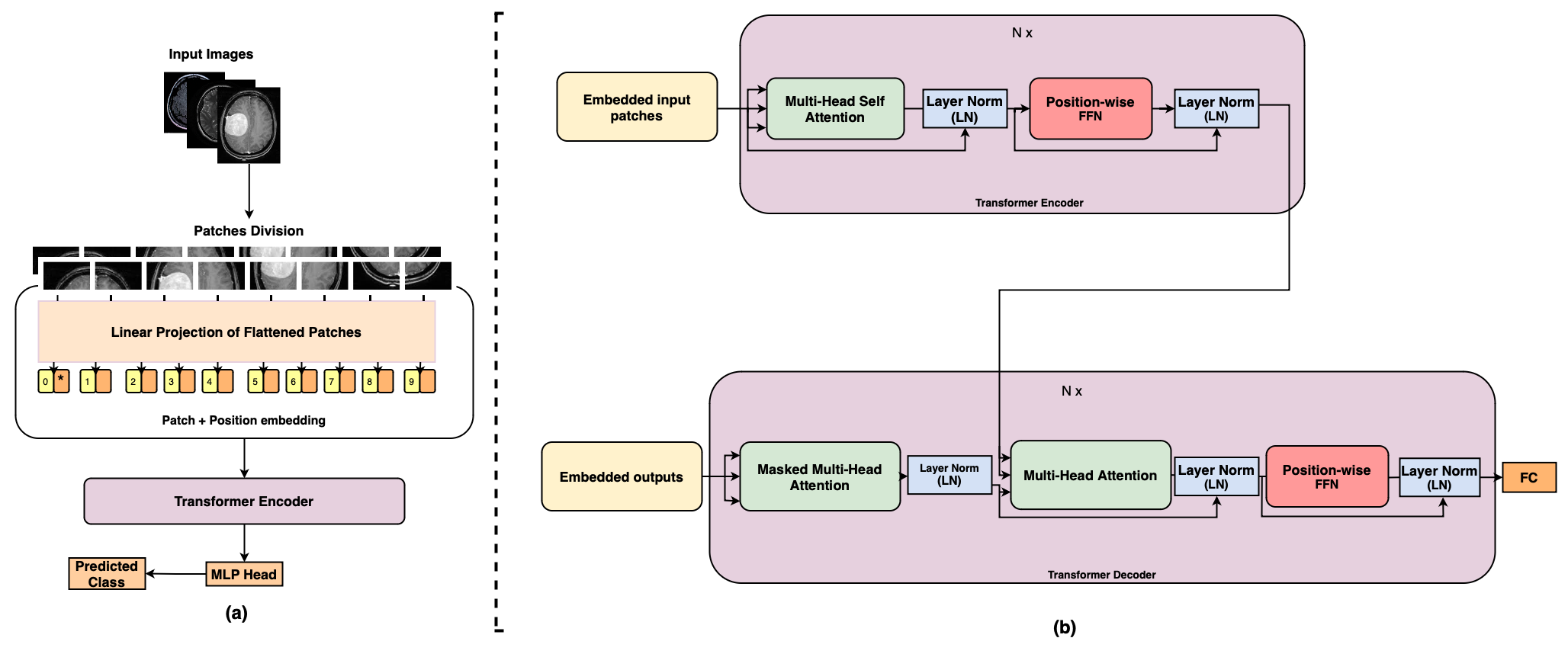}
\caption{(a):  An image is divided into fixed-size patches, each is embedded linearly, and position embeddings are added. The sequence of vectors produced is then fed into a standard Transformer encoder. For classification purposes, an additional learnable classification token is incorporated into the sequence. (b): The Transformer's architecture is characterized by the use of stacked self-attention and point-wise, fully connected layers within both its encoder and decoder components, as depicted in the left and right sections of the figure, respectively.}
\label{fig_structure}
\end{figure*}

The transformer model, initially applied in the field of natural language processing (NLP) for machine translation tasks \cite{vaswani2017attention}, consists of an encoder and a decoder. Both the encoder and the decoder are composed of multiple transformer blocks, each having an identical architecture. Figure \ref{fig_structure} illustrates the basic configuration of ViTs. The encoder is responsible for generating encodings of the input. In contrast, the decoder leverages the contextual information embedded within these encodings to generate the output sequence. Each transformer block within the model encompasses several components: a multi-head attention layer, a feed-forward neural network, residual connections, and layer normalization.  \\

\subsection{Overview of the Vision Transformers Architecture}
The advancements in basic transformer models are largely due to their two main components. The first is the self-attention mechanism, which excels in capturing long-range dependencies among sequence elements. This surpasses the limitations of traditional recurrent models in encoding such relationships. The second key component is the transformer encoder layers. These layers are pivotal in hierarchical representation learning within transformer models, as they integrate self-attention with feed-forward networks. This integration enables effective feature extraction and information propagation throughout the model \cite{khan2022transformers, lin2022survey, dosovitskiy2020image}.

\textbf{Self-attention mechanism:}
The self-attention mechanism assesses the importance or relevance of each patch in a sequence in relation to others. For instance, in language processing, it can identify words that are likely to co-occur in a sentence. As a fundamental part of transformers, self-attention captures the interactions among all elements in a sequence, which is especially beneficial for tasks that involve structured predictions \cite{khan2022transformers}. A self-attention layer updates each sequence element by aggregating information from the entire input sequence.\\

Let's denote a sequence of $N$ entities $(x_1,x_2, \dots ,x_n)$ by $\mathbf{X} \in {R}^{n \times d}$, where $d$ is the embedding dimension for each entity. The goal of self-attention is to capture the relationships among all the entities by encoding each entity based on the overall contextual information. To achieve this, it employs three learnable weight matrices: Queries $\left(\mathbf{W}^Q \in {R}^{d \times d_q}\right)$, Keys $\left(\mathbf{W}^K \in {R}^{d \times d_k}\right)$, and Values $\left(\mathbf{W}^V \in {R}^{d \times d_v}\right)$, where $d_q$=$d_k$. By projecting the input sequence $X$ onto these weight matrices, it obtains $Q = X \cdot W^Q$, $K = X \cdot W^K$, and $V= X \cdot W^V$. The self-attention layer outputs $Z \in {R}^{n \times {d_v}}$ \cite{vaswani2017attention}, calculated as:
\begin{equation}
     \mathbf{Z}=\operatorname{softmax}\left(\frac{\mathbf{Q K}^T}{\sqrt{d_q}}\right) \mathbf \cdot V
     \end{equation}
\\
To determine the importance or weight of each value in the sequence, a softmax function is applied. This function assigns weights to the values based on their relevance or significance within the context of the task or model.
In summary, the self-attention mechanism enables each element in the sequence to be updated based on its interactions with other elements, incorporating global contextual information. The self-attention mechanism computes the dot product between the query and all the keys for a specific entity in the sequence. These dot products are then normalized using the softmax function, resulting in attention scores. Each entity in the sequence is then updated as a weighted sum of all the entities, with the weights determined by the attention scores. We will delve deeper into scaled dot-product attention and its application within multi-head attention in the following section. \ref{multi_head}.\\

\textbf{Transformer encoder and decoder layers:}
The encoder is composed of a sequence of identical layers, with a total of $N$ layers, where $N$ is specified in Figure \ref{fig_structure}. Each layer comprises two principal sub-layers: a multi-head self-attention mechanism and a position-wise fully connected feed-forward network. Subsequent to each layer, the architecture employs residual connections \cite{he2016deep} and layer normalization \cite{ba2016layer}. This configuration stands in contrast to CNNs, in which feature aggregation and transformation are executed concurrently. Within the transformer architecture, these operations are distinctly partitioned: the self-attention sub-layer is tasked with aggregation exclusively, whereas the feed-forward sub-layer focuses on the transformation of features.

The decoder is structured similarly, consisting of identical layers. Each layer within the decoder encompasses three sub-layers. The initial two sub-layers, specifically the multi-head self-attention and the feed-forward networks, reflect the architecture of the encoder. The third sub-layer introduces a novel multi-head attention mechanism that targets the outputs from the corresponding encoder layer, as depicted in Figure \ref{fig_structure}-b \cite{khan2022transformers}.

\subsection{Key Components and Building Blocks of Vision Transformers}
The subsequent sections will provide in-depth explanations of the key components and fundamental building blocks of ViTs. These include patch extraction and embedding, positional encoding, multi-head self-attention, and feed-forward networks. In patch extraction, an image is divided into smaller patches, each of which is then transformed into a numerical representation through an embedding process. Positional encoding is employed to incorporate spatial information, allowing the model to account for the relative positions of these patches. The multi-head self-attention mechanism is crucial for capturing dependencies and contextual relationships within the image. Finally, the feed-forward networks are responsible for introducing non-linear transformations, enhancing the model's ability to process complex visual information.\\

\textbf{Patch extraction and embedding:}
A pure transformer model can be directly employed for image classification tasks by operating on sequences of image patches. This approach adheres closely to the original design of the transformer. To handle $2D$ images, the input image $X \in \mathbb{R}^{h \times w \times c}$ is reshaped into a sequence of flattened 2D patches $X_p \in \mathbb{R}^{n \times (p^2 \times c)}$, where $c$ represents the number of channels. The original image resolution is denoted as $(h, w)$, while $(p, p)$ signifies the resolution of each image patch. The effective sequence length for the transformer is defined as $n = \frac{h \times w}{p^2}$. Given that the transformer employs consistent dimensions across its layers, a trainable linear projection is applied to map each vectorized patch to the model dimension 
$d$. This output is referred to as patch embedding \cite{dosovitskiy2020image, han2022survey}.

\textbf{Positional encoding:}
To optimize the model's use of sequence order, integrating information about the tokens' relative or absolute positions is essential. This is accomplished through the addition of positional encoding to the input embeddings at the foundation of both the encoder and decoder stacks. These positional encodings, matching the dimensionality $d_{\text{model}}$ of the embeddings, are merged with the input embeddings. Positional encoding can be generated through various methods, including both learned and fixed strategies \cite{gehring2017convolutional}. The precise technique for embedding positional information is delineated by the equations that follow:\\
\begin{equation}
   \text{PE}(\text{pos}, 2i) = \sin\left(\frac{\text{pos}}{10000^{(2i/d_{\text{model}})}}\right)
\end{equation}\\

\begin{equation}  
   \text{PE}(\text{pos}, 2i+1) = \cos\left(\frac{\text{pos}}{10000^{(2i/d_{\text{model}})}}\right) 
\end{equation}

\noindent In the given equations, $pos$ represents the position of a word within a sentence, and $i$ refers to the current dimension of the positional encoding. In this manner, the positional encoding in the transformer model assigns a sinusoidal value to each element. This enables the model to learn relative positional relationships and generalize to longer sequences during inference. In addition to the fixed positional encoding employed in the original transformer, other models have explored the use of learned positional encoding \cite{gehring2017convolutional} and relative positional encoding \cite{shaw2018self, devlin2019google, dosovitskiy2020image}.

\begin{figure}[h]
    \centering \includegraphics[width=0.9\linewidth]{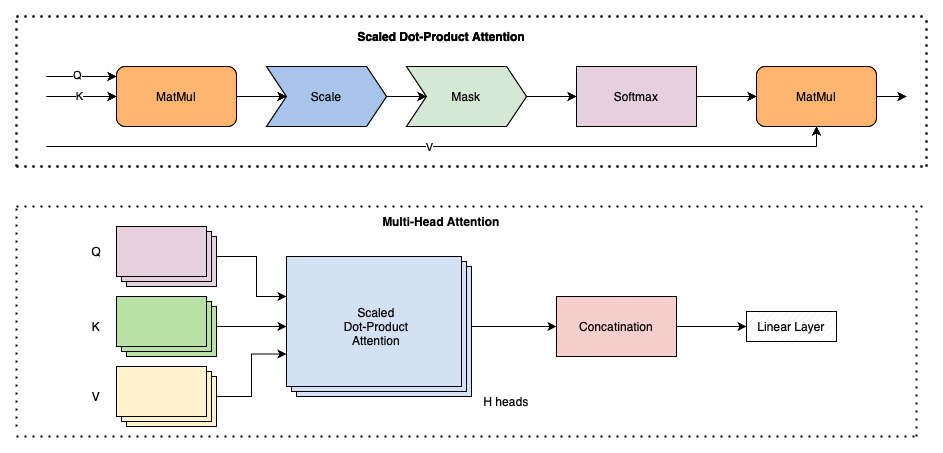} 
    \caption{Schematic representation of the Scaled Dot-Product Attention and Multi-Head Attention mechanisms. The top process combines queries, keys, and values to compute attention scores, while the bottom shows parallel attention layers merging in Multi-Head Attention, a core feature of transformer models for capturing varied contextual cues. The depiction of the attention mechanism inspired by \cite{khan2022transformers}.}
    \label{Scaled dot Product_Multi head attention}
\end{figure}

\textbf{Multi-head self-attention:}
\label{multi_head}
The multi-head attention mechanism enables the model to capture multiple complex relationships among different elements in the sequence. It achieves this by utilizing several self-attention blocks, each with its own set of weight matrices. The outputs of these blocks are combined and projected onto a weight matrix to obtain a comprehensive representation of the input sequence.
The original transformer model employs h = 8 blocks. Each block has its own distinct set of learnable weight matrices, denoted as ${W^{Q_i}, W^{K_i}, W^{V_i}}$ for $ i = 0, 1, ..., h-1$. When given an input X, the output of the h self-attention blocks in the multi-head attention is concatenated into a single matrix $\left(\mathbf{[Z_0, Z_1, \dots, Z_h-1]} \in {R}^{n \times{h . d_v}}\right)$. This concatenated matrix is then projected on to a weight matrix $\left(\mathbf{W} \in {R}^{{h . d_v}\times d}\right)$ \cite{khan2022transformers}.  Refer to Figure \ref{Scaled dot Product_Multi head attention} for a visual overview of the Scaled Dot-Product Attention and its extension into Multi-Head Attention, fundamental mechanisms for contextual processing in transformer architectures. The diagram details the flow from input queries to the final attention output.

Self-attention differs from convolutional operations in that it calculates filters dynamically rather than relying on static filters. Unlike convolution, self-attention is invariant to permutations and changes in the number of input points, allowing it to handle irregular inputs effectively. It has been shown in research that self-attention, when used with positional encodings, offers greater flexibility and can effectively capture local features similar to convolutional models \cite{perez2019turing, cordonnier2019relationship}. Further investigations have been conducted to analyze the relationship between self-attention and convolution operations. Empirical evidence supports the notion that multi-head self-attention, with sufficient parameters, serves as a more general operation that can encompass the expressiveness of convolution. In fact, self-attention possesses the capability to learn both global and local features, enabling it to adaptively determine kernel weights and adjust the receptive field, similar to deformable convolutions. This demonstrates the versatility and effectiveness of self-attention in capturing diverse aspects of data \cite{dai2017deformable}.

\textbf{Feed-forward networks:}
In both the encoder and decoder, a feed-forward network (FFN) follows the self-attention layers. This network incorporates two linear transformation layers and a nonlinear activation function within them. Denoted as the function $FFN(X)$, it can be expressed as $FFN(X) = W_2S(W_1X)$, where $W_1$ and $W_2$ are the parameter matrices of the linear transformation layers, and $S$ represents the chosen nonlinear activation function, such as GELU \cite{hendrycks2016gaussian}.

\subsection{Training process of Vision Transformers}
Self-attention-based transformer models have revolutionized machine learning through their extensive pre-training on large datasets. This pre-training stage employs a variety of learning approaches, including supervised, unsupervised, and self-supervised methods. Such methodologies have been explored in the seminal works of Dosovitskiy et al. \cite{dosovitskiy2020image}, Devlin et al. \cite{devlin2018bert}, Li et al. \cite{li2020oscar}, and Lin et al. \cite{lin2021end}. The primary goal of this phase is to acclimate the model to a broad spectrum of data, or to a combination of different datasets. This strategy aims to establish a foundational understanding of visual information processing, a concept further elucidated by Su et al. \cite{su2019vl} and Chen et al. \cite{chen2020uniter}.

After pre-training, these models undergo fine-tuning with more specialized datasets, which vary in size. This step is crucial for tailoring the model to specific applications, such as image classification \cite{gidaris2018unsupervised}, object detection \cite{carion2020end}, and action recognition \cite{gidaris2018unsupervised}, thereby improving their performance and accuracy in these tasks.

The value of pre-training is particularly evident in large-scale transformer models utilized across both language and vision domains. For instance, the Vision Transformer (ViT) model exhibits a marked decline in performance when trained exclusively on the ImageNet dataset, as opposed to including pre-training on the more comprehensive JFT-300M dataset, which boasts over 300 million images \cite{dosovitskiy2020image, gupta2017revisiting}. While pre-training on such extensive datasets significantly boosts model performance, it introduces a practical challenge: manually labeling vast datasets is both labor-intensive and costly.

This challenge leads researchers to the pivotal role of self-supervised learning (SSL) in developing scalable and efficient transformer models. SSL emerges as an effective strategy by using unlabeled data, thereby avoiding the limitations associated with extensive manual annotation. Through SSL, models undertake pretext tasks that generate pseudo-labels from the data itself, fostering a foundational understanding of data patterns and features without the necessity for explicit labeling \cite{jing2020self, liu2021self}. This method not only enhances the model's ability to discern crucial features and patterns, pivotal for downstream tasks with limited labeled data but also maximizes the utility of the vast volumes of unlabeled data available.

Contrastive learning, a subset of SSL, exemplifies this by focusing on identifying minor semantic differences in images, which significantly sharpens the model's semantic discernment \cite{khan2022transformers}. The transition from traditional pre-training methods to SSL underscores a paradigm shift in how models are trained, moving from reliance on extensive, manually labeled datasets to an innovative use of unlabeled data. This shift not only addresses the scalability and resource challenges but also enhances the generalizability and efficiency of transformer networks.

Khan et al. \cite{khan2022transformers} categorize SSL methods based on their pretext tasks into generative, context-based, and cross-modal approaches. Generative methods focus on creating images or videos that match the original data distribution, teaching the model to recognize data patterns. Context-based methods use spatial or temporal relationships within the data, enhancing contextual understanding. Cross-modal methods exploit correspondences between different data types, like image-text or audio-video, for a more comprehensive data understanding.

Generative approaches, especially those involving masked image modeling, train models to reconstruct missing or obscured parts of images, thereby refining their generative skills \cite{chen2020uniter}. Other SSL strategies include image colorization \cite{pathak2016context}, image super-resolution \cite{ledig2017photo}, image in-painting \cite{pathak2016context}, and approaches using GAN networks \cite{goodfellow2014generative, alijani2022ensemble}. Context-based SSL approaches deal with tasks like solving image patch jigsaw puzzles \cite{ahsan2019video}, classifying masked objects \cite{su2019vl}, predicting geometric transformations like rotations \cite{gidaris2018unsupervised}, and verifying the chronological sequence of video frames \cite{lee2017unsupervised}. Finally, cross-modal SSL methods focus on aligning different modalities, ensuring correspondences between elements such as text and image \cite{li2019visualbert}, audio and video \cite{korbar2018cooperative}, or RGB and flow information \cite{sayed2019cross}.

\subsection{Advantages of Vision Transformers compared to CNNs backbones}
The advent of ViTs represents a significant innovation in the field of image processing, offering substantial benefits over traditional CNNs like ResNet. These advantages are succinctly demonstrated which underscores the key distinctions and superiorities of ViTs:

\textbf{Performance Improvement:} ViTs have been effectively adapted for a broad range of vision recognition tasks, demonstrating significant enhancements over CNNs. These advancements are particularly pronounced in tasks such as classification on the ImageNet dataset \cite{dosovitskiy2020image}, object detection \cite{carion2020end}, and semantic segmentation \cite{liu2021swin}, areas where ViTs have outperformed established benchmarks. Remarkably, ViTs have also shown the capability to achieve competitive results with architectures that are smaller in size, highlighting their efficiency and scalability \cite{ranftl2021vision}. Furthermore, the overall improvements brought by ViTs in the vision domain are supported by comprehensive analyses and comparisons \cite{khan2022transformers}.

\textbf{Exploiting Long-Range Dependencies, the Power of Attention Mechanisms in ViTs:} The attention mechanism within ViTs effectively captures long-range dependencies in the input data. This modeling of inter-token relationships facilitates a more comprehensive global context, representing a significant advancement beyond the local processing capabilities of CNNs \cite{shao2021adversarial, touvron2021training, matsoukas2021time} and more recent works \cite{li2024efficient}. Additionally, the attention mechanism provides insight into the focus areas of the model during input processing, acting as a built-in saliency map \cite{caron2021emerging}. 

\textbf{Flexibility and Extensibility:} ViTs have proven to be highly versatile, serving as a backbone that surpasses previous benchmarks with their dynamic inference capabilities. They are particularly adept at handling unordered and unstructured point sets, making them suitable for a broader range of applications \cite{doersch2020crosstransformers, zhao2021point}.

\textbf{Enhanced Text-Visual Integration with ViTs:} The ability of ViTs to integrate text and visual data facilitates an unparalleled understanding of the dependencies between different tasks, effectively harnessing the synergy between diverse data types \cite{lu2019vilbert, plummer2015flickr30k}. Although CNNs can be adapted for text-visual fusion—potentially with the assistance of Recurrent Neural Networks (RNNs), ViTs excel in this area. This superiority stems from ViTs' inherent design, which naturally accommodates the parallel processing of complex, multimodal datasets. Unlike CNNs, which may require additional mechanisms or complex architectures to achieve similar integrations, ViTs directly leverage their attention mechanisms to dynamically weigh the importance of different data elements.

\textbf{End-to-End Training:} The architecture of ViTs facilitates end-to-end training for tasks such as object detection, streamlining the training process by obviating the need for complex post-processing steps \cite{carion2020end}.

In essence, ViTs offer a powerful modeling approach that is well-suited to extracting meaningful information from vast and varied input data. These visualizations depict how each network type processes and identifies areas of focus within the image. The attention maps from the ViT show distinct patterns indicating specific regions the model attends to when making predictions, while the feature maps from the ResNet indicate more dispersed and convolutionally derived features throughout the image.

\section{Vision Transformers in Domain Adaptation and Domain Generalization}
\label{sec:3}
ViTs have shown promising performance in computer vision tasks. In this section, we focus on exploring the potential of adapting ViT to DA and DG scenarios to mitigate the distribution shifts.

Recent studies have demonstrated that leveraging ViTs backbones as the feature extractors offers superior capability in managing distribution shifts compared to conventional CNN architectures \cite{xu2021cdtrans, yang2023tvt, sun2022safe}. This superiority is relevant for practical applications where adapting to varied data distributions is important. These works have led researchers to develop multi-modal approaches in DA and DG scenarios, integrating diverse strategies to further enhance the adaptability and generalization capabilities of ViTs. These approaches often involve the strategic selection of different models and integrating them and carefully chosen loss functions, aimed at regularizing the training of multi-modals designs \cite{angarano2022back, yang2023tvt, sun2022safe, ma2022making, zhu2023patch}.

\begin{figure}[h]
    \centering \includegraphics[width=0.9\linewidth]{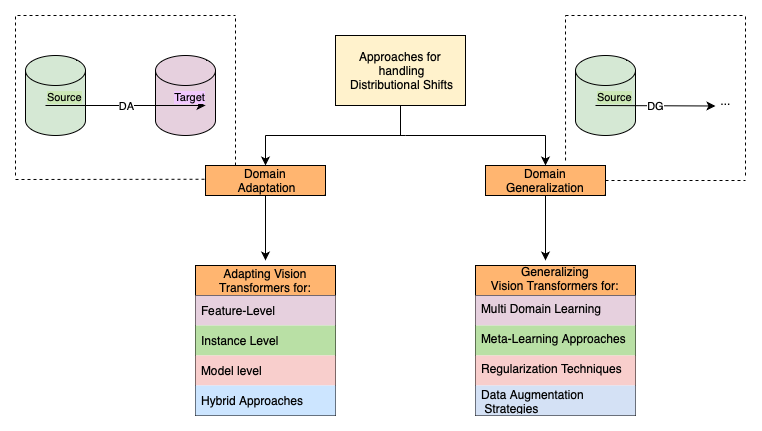} 
    \caption{Our categorization of studies on adapting vision transformers to handle distribution shifts in domain adaptation and domain generalization approaches.}
    \label{ViTs For DA/DG}
\end{figure}

In the following sections of our paper, we aim to provide an in-depth analysis of these methodologies and their implications in the field of computer vision. This will include detailed discussions on the specific techniques and innovations that have enhanced the performance of ViTs in regard to distribution shifts. Figure \ref{ViTs For DA/DG} illustrates the categorization of the research we have reviewed for this paper, highlighting the respective approaches within DA and DG methods. In the upcoming sections, we aim to provide a more comprehensive explanation of the methods employed.

\subsection{Vision Transformers in Domain Adaptation}
DA is a critical area of research within machine learning that aims to improve model performance on a target domain by leveraging knowledge from a source domain, especially when the data distribution differs between these domains. This discrepancy, known as a distribution shift, poses significant challenges to the adapting of models across varied application scenarios. DA techniques are designed to mitigate these challenges by adapting models to perform well on data that was not seen during training, thereby enhancing model robustness and generalizability \cite{wang2018deep}.

Utilizing ViTs to address distribution shifts within the framework of DA approaches., offer novel methods to model robustness and generalizability in diverse application scenarios. In the majority of studies exploring the application of ViTs to address the challenges of distribution shifts within DA strategies, the primary emphasis has been on unsupervised domain adaptation (UDA). DA strategies in the context of ViTs can be broadly classified into several categories, each contributing uniquely to the model's adaptability to different target domains. Our categorization are based on feature-level adaptation, instance-level adaptation, model-level adaptation, and hybrid approaches. In each of these categories, we further elaborate on the adaptation level, providing insights into their architecture, and efficacy in DA scenarios.\\

\subsubsection{Feature-Level Adaptation} Feature-level adaptation involves aligning the feature distributions between the source and target domains to ensure that the features learned by the model in the source domain are applicable to the target domain. This approach is particularly effective in addressing the domain shift problem by transforming the feature space of the source domain to closely match that of the target domain. Techniques such as Domain-Oriented Transformer (DOT), TRANS-DA, and Spectral UDA (SUDA) have shown promising results by employing various strategies like adversarial training, feature matching, and domain-specific normalization. By aligning feature distributions, these methods help to reduce the discrepancy between domains, thereby improving the model's performance on the target domain.\\
Researchers proposed a Domain-Oriented Transformer to address the challenges faced by conventional UDA techniques for domain discrepancies. Traditional methods often encounter difficulties when attempting to align domains, which can compromise the discriminability of the target domain when classifiers are biased towards the source data. To overcome these limitations, DOT employs feature alignment across two distinct spaces, each specifically designed for one of the domains. It leverages separate classification tokens and classifiers for each domain. This approach ensures the preservation of domain-specific discriminability while effectively capturing both domain-invariant and domain-specific information. It achieves this through a combination of contrastive-based alignment and source-guided pseudo-label. The novel DOT method is introduced in \cite{ma2022making}. 

TRANS-DA \cite{ye2023learning}, focuses on generating pseudo-labels with reduced noise and retraining the model using new images composed of patches from both source and target domains. This approach includes a cross-domain alignment loss for better matching centroids of labeled and pseudo-labeled patches, aiming to improve domain adaptation. It falls into the feature-level adaptation category, as it focuses on refining feature representations and aligning them across domains.
Another study shifts focus to integrating transformers with CNN-backbones, proposing Domain-Transformer for UDA. This approach, distinct from existing methods that rely heavily on local interactions among image patches, introduces a plug-and-play domain-level attention mechanism. This mechanism emphasizes transferable features by ensuring local semantic consistency across domains, leveraging domain-level attention and manifold regularization \cite{chuan2022towards}.

Spectral UDA (SUDA) \cite{zhang2022spectral}, is an innovative UDA technique in spectral space. SUDA introduces a Spectrum Transformer (ST) for mitigating inter-domain discrepancies and a multi-view spectral learning approach for learning diverse target representations. The paper's approach emphasizes feature-level adaptation, focusing on learning domain-invariant spectral features efficiently and effectively across various visual tasks such as image classification, segmentation, and object detection.
In \cite{li2022semantic} the Semantic Aware Message Broadcasting (SAMB) is introduced to enhance feature alignment in UDA. This approach challenges the effectiveness of using just one global class token in ViTs. It suggests adding group tokens instead. These tokens focus on broadcasting messages to different semantic regions, thereby enriching domain alignment features. Additionally, the study explores the impact of adversarial-based feature alignment and pseudo-label based self-training on UDA, proposing a two-stage training strategy that enhances the adaptation capability of the ViT \cite{li2022semantic}.

Gao et al. in \cite{gao2022visual} addresses the Test Time Adaptation (TTA) challenge for adapting to target data and avoiding performance degradation due to distribution shifts. By introducing Data-efficient Prompt Tuning (DePT), the approach combines visual prompts in ViTs with source-initialized prompt fine-tuning. This fine-tuning, paired with a memory bank-based online pseudo-labeling and hierarchical self-supervised regularization, enables efficient model adjustment to the target domain, even with minimal data. DePT's adaptability extends to online or multi-source TTA settings \cite{gao2022visual}. Furthermore, CTTA \cite{gan2023decorate} proposes a unique approach to continual TTA using visual domain prompts. It presents a lightweight, image-level adaptation strategy, where visual prompts are dynamically added to input images, adapting them to the source domain model. This approach mitigates error accumulation and catastrophic forgetting by focusing on input modification rather than model tuning, a significant shift from traditional model-dependent methods. This method can be classified as a feature-level adaptation, as it primarily focuses on adjusting the input image features for domain adaptation without altering the underlying model architecture.

\subsubsection{Instance-Level Adaptation} Instance-level adaptation involves selecting or weighting specific data points (instances) more heavily during training to ensure that the model learns features relevant to the target domain. This approach can significantly improve the model's performance on the target domain by prioritizing instances that reflect the characteristics of the target data. Techniques such as Source Free Open Set Domain Adaptation (SF-OSDA), style-based data augmentation, and clustering are commonly used in instance-level adaptation to refine and enhance the training process. By focusing on relevant instances, these methods help to reduce the impact of domain shift and improve the model’s generalization capabilities.

One notable instance of instance-level adaptation is addressed in the study by \cite{vray2023source}, which deals with the challenge of Source Free Open Set Domain Adaptation. This scenario involves adapting a pre-trained model, initially trained on an inaccessible source dataset, to an unlabeled target dataset that includes open set samples, or data points that do not belong to any class seen during training. The primary technique involves leveraging a self-supervised ViT, which learns directly from the target domain to distill knowledge. A crucial element of their method is a unique style-based data augmentation technique, designed to enhance the training of the ViT within the target domain by providing a richer, more contextually diverse set of training data. This leads to the creation of embeddings with rich contextual information.

The model uses these information-rich embeddings to cluster target images based on semantic similarities and assigns them weak pseudo-labels with associated uncertainty levels. To improve the accuracy of these pseudo-labels, the researchers introduce a metric called Cluster Relative Maximum Logit Score (CRMLS). This measure adjusts the confidence levels of the pseudo-labels, making them more reliable. Additionally, the approach calculates weighted class prototypes within this enriched embedding space, facilitating the effective adaptation of the source model to the target domain, thus exemplifying an application of instance-level adaptation techniques.

\subsubsection{Model-Level Adaptation}
 Model-level adaptation involves developing specialized ViT architectures or modifying existing models to enhance their adaptability to domain shifts. This approach focuses on adapting the internal structure of the model itself to better handle variations between the source and target domains. Techniques such as introducing new layers, modifying attention mechanisms, and designing domain-specific model components are commonly employed in model-level adaptation. These modifications enable the model to learn more robust and transferable features, improving its performance on the target domain.

Zhang et al. in \cite{zhang2022delving} primarily focus on enhancing the out-of-distribution generalization of ViTs. It delves into techniques like adversarial learning, information theory, and self-supervised learning to improve model robustness against distribution shifts. The study is categorized into model-level adaptation, as it enhances the general model architecture and training process of ViTs to achieve better performance across varied distributions. Yang et al \cite{yang2021transformer} introduce TransDA, a novel framework for source-free domain adaptation (SFDA), which integrates a transformer with a CNN to enhance focus on important object regions. Diverging from traditional SFDA approaches that primarily align cross-domain distributions, TransDA capitalizes on the initial influence of pre-trained source models on target outputs. By embedding the transformer as an attention module in the CNN, the model gains improved generalization capabilities for target domains. Additionally, the framework employs self-supervised knowledge distillation using target pseudo-labels to refine the transformer's attention towards object regions. This approach effectively addresses the limitations of CNNs in handling significant domain shifts, which often lead to over-fitting and a lack of focus on relevant objects, therefore offering a more robust solution for domain adaptation challenges. In addressing the intricacies of model-level adaptation in DA, a series of innovative approaches emerge, each offering unique solutions to prevalent challenges. \cite{tayyab2021pre} highlights the use of BeiT, a pre-trained transformer model, for UDA. The core idea is leveraging BeiT's powerful feature extraction capabilities, initially trained on source datasets, and then adapting them to target datasets. This approach, which significantly outperforms existing methods in the ViSDA Challenge, primarily focuses on model-level adaptation. It utilizes the self-attention mechanism inherent in transformers to adapt to new, out-of-distribution target datasets, demonstrating a marked improvement in domain adaptation tasks. TFC \cite{wang2022tfc} aims to bridge this gap by demonstrating the potential of combining convolutional operations and transformer mechanisms for adversarial UDA through a hybrid network structure termed transformer fused convolution (TFC). By seamlessly integrating local and global features, TFC enhances the representation capacity for UDA and improves the differentiation between foreground and background elements. Additionally, to bolster TFC's resilience, an uncertainty penalty loss is introduced, leading to the consistent assignment of lower scores to incorrect classes.

\subsubsection{Hybrid Approaches}
In our categorization of domain adaptation techniques, hybrid approaches combine multiple methods (feature-level, instance-level, and model-level) to leverage the strengths of each. By integrating various techniques, hybrid approaches aim to provide a more robust and comprehensive solution to domain adaptation challenges. These methods effectively address the limitations of individual adaptation strategies by simultaneously aligning feature distributions, selecting relevant instances, and modifying model architectures. Recent studies that fall into our hybrid approaches category include Cross-Domain Vision Transformer (CDTrans), Augmented Transformer, and Multi-View Adaptation. These studies illustrate the effectiveness of hybrid approaches in improving the performance of ViTs across diverse domains.

CDTRANS \cite{xu2021cdtrans} introduces a hybrid domain adaptation approach via a triple-branch transformer that combines feature-level and model-level adaptations. It incorporates a novel cross-attention module along with self-attention mechanisms within its architecture. The design includes separate branches for source and target data processing and a third for aligning features from both domains. This setup enables simultaneous learning of domain-specific and domain-invariant representations, showing resilience against label noise. Additionally, the paper proposes a two-way center-aware labeling algorithm for the target domain, utilizing a cross-domain similarity matrix to enhance pseudo-label accuracy and mitigate noise impact. This approach effectively merges feature acquisition with alignment, showcasing a sophisticated method for handling domain adaptation challenges.
TVT (Transferable Vision Transformer) \cite{yang2023tvt} explores the use of ViT in UDA. It examines ViT's transferability compared to CNNs and proposes the TVT framework, which includes a Transferability Adaptation Module (TAM) and a Discriminative Clustering Module (DCM). TVT integrates various domain adaptation approaches. It employs model-level adaptations by leveraging ViTs, and optimizing them for UDA. Simultaneously, it involves feature-level adaptation through the TAM, which enhances feature representations for better alignment between source and target domains. Furthermore, instance-level strategies are incorporated via the Discriminative Clustering Module (DCM), focusing on the diversification and discrimination of patch-level features. This multifaceted approach, combining model, feature, and instance-level adaptations, exemplifies a hybrid strategy in domain adaptation.

SSRT \cite{sun2022safe} enhances domain adaptation by integrating a ViT backbone with a self-refinement strategy using perturbed target domain data. The approach includes a safe training mechanism that adaptively adjusts learning configurations to avoid model collapse, especially in scenarios with large domain gaps. This novel solution showcases both model-level adaptations, by employing ViTs, and feature-level adaptations, through its unique self-refinement method. BCAT \cite{wang2022domain} presents a novel framework which introduces a bidirectional cross-attention mechanism to enhance the transferability of ViTs. This mechanism focuses on blending source and target domain features to minimize domain discrepancy effectively. The BCAT model combines this with self-attention in a unique quadruple transformer block structure to focus on both intra and inter-domain features. It has a novel transformer architecture with a bidirectional cross-attention mechanism for model-level adaptation, and integration and alignment of features from different domains for the instance-level adaptation.

 PMTrans \cite{zhu2023patch} employs a PatchMix transformer to bridge source and target domains via an intermediate domain, enhancing domain alignment. It conceptualizes UDA as a min-max cross-entropy game involving three entities: feature extractor, classifier, and PatchMix module. This unique approach, leveraging a mix of patches from both domains, leads to significant performance gains on benchmark datasets. The paper aligns with hybrid adaptation, combining model-level innovations (PatchMix Transformer) and feature-level strategies (patch mixing for domain bridging). UniAM \cite{zhu2023universal}, for Universal DA (UniDA) leverages ViTs and introduces a Compressive Attention Matching (CAM) approach to address the UniDA problem. This method focuses on the discriminability of attention across different classes, utilizing both feature and attention information. UniAM stands out by its ability to effectively handle attention mismatches and enhance common feature alignment. The paper fits into the hybrid category of domain adaptation, combining feature-level strategies with model-level adaptations through the use of ViT and attention mechanisms.
 
 CoNMix \cite{kumar2023conmix} presents a novel framework for source-free DA, adept at tackling both single and multi-target DA challenges in scenarios where labeled source data is unavailable during target adaptation. This framework employs a Vision ViT as its backbone and introduces a distinctive strategy that integrates consistency with two advanced techniques: Nuclear-Norm Maximization and MixUp knowledge distillation. Nuclear-Norm Maximization is a regularization technique that encourages the model to learn a low-rank representation of data, promoting generalization by reducing complexity. MixUp knowledge distillation, on the other hand, leverages a data augmentation method that combines inputs and labels in a weighted manner to create synthetic training examples, enhancing the model's ability to generalize across domains. The framework demonstrates state-of-the-art results across various domain adaptation settings, showcasing its effectiveness in scenarios with privacy-related restrictions on data sharing. This paper aligns with a hybrid adaptation approach, incorporating both model-level and feature-level (consistency and pseudo-label refinement methods) strategies.
 
 \cite{ma2021exploiting} introduces the Win-Win Transformer (WinTR) framework. This framework effectively leverages dual classification tokens in a transformer to separately explore domain-specific knowledge for each domain while also interchanging cross-domain knowledge. It incorporates domain-specific classifiers for each token, emphasizing the preservation of domain-specific information and facilitating knowledge transfer. This approach exhibits significant performance improvements in UDA tasks. The paper exemplifies a hybrid approach, combining model-level adaptations with its transformer structure and feature-level strategies through domain-specific learning and knowledge transfer mechanisms.
 
 Table \ref{tab:Design_highlights_ViT} provides a comprehensive summary of the main categories in adapting ViT for DA: feature level adaptation, instance level adaptation, model level adaptation, and hybrid approaches. It details the various methods used, design highlights, and different loss functions employed during training. Additionally, the table references the publication details for each representative study, including the journals or conferences where they were published.


\begin{sidewaystable}
\caption{Representative Works of ViTs for DA}\label{tab:Design_highlights_ViT}
\begin{tabularx}{\textheight}{L{2cm} L{2cm} Y L{3cm} L{2cm}}
\toprule
\makecell{Category} & \makecell{Method} & \makecell{Design Highlights} & \makecell{Loss Functions} & \makecell{Publication} \\
\midrule
Feature-level Adaptation & DOT \cite{ma2022making} & Domain-Oriented Transformer with dual classifiers utilizes individual tokens for source/target domain adaptation and domain-specific learning, enhanced by source-guided pseudo-label refinement & Cross-entropy, Contrastive, Representation Difference & ICMR 2022 \\
\\

& SUDA \cite{zhang2022spectral} & Enhancing visual task generalization, leveraging Spectrum Transformer for domain alignment and multi-view learning to optimize mutual information & Supervised, Discrepancy, Unsupervised Similarity & CVPR 2022 \\
\hline
Model-level Adaptation & GE-ViT \cite{zhang2022delving} & Leveraging ViTs' inductive biases towards shapes and structures, combined with adversarial and self-supervised learning techniques & Cross-Entropy, Entropy, Adversarial & CVPR 2022 \\
\\
& TFC \cite{wang2022tfc} & Integrates CNN and ViTs for both capturing of local and global features, achieving distinct foreground-background separation, further refined by an uncertainty-based accuracy enhancement & Cross-Entropy, Adversarial, Uncertainty Penalization & IEEE-TCSS 2022 \\
\hline
Hybrid Approaches & CDTRANS \cite{xu2021cdtrans} & Triple-branch Transformer with shared weights aligns source/target data, using self/cross-attention and center-aware labeling for enhanced pseudo-label accuracy and noise distillation & Cross-entropy, Distillation & ICLR 2022 \\
\\
& TVT \cite{yang2023tvt} & Enhances feature learning via adversarial adaptation, using transferability and clustering modules for robust knowledge transfer, and fine-tunes attention for precise domain discrimination & Cross-Entropy, Adversarial, Patch-Level Discriminator & WACV 2023 \\
\\
& SSRT \cite{sun2022safe} & Utilizes adversarial adaptation and enhances model accuracy by refining through perturbed data feedback, incorporating a robust training approach to minimize KL divergence & Cross-Entropy, Self-Refinement, Domain Adversarial & CVPR 2022 \\
\\
& PMTrans \cite{zhu2023patch} & Employs a module based on transformer architecture for domain representation, leveraging game theory for patch sampling, and combines source/target data to optimize label accuracy with attention-driven adjustments & Cross-Entropy, Semi-Supervised Mixup & CVPR 2023 \\
\\
& SMTDA \cite{kumar2023conmix} & Enhances domain adaptation by maximizing nuclear-norm consistency and distilling knowledge via MixUp, refining pseudo labels for broadened generalization across multiple targets & Cross-Entropy, Nuclear-Norm, Knowledge Distillation & WACV 2023 \\
\\
& UniAM \cite{zhu2023universal} & Utilizes universal and compressive attention for precise alignment and class separation, achieving categorization without prior label knowledge & Cross-Entropy, Adversarial, Source-Target Contrastive & ICCV 2023 \\
\hline
\end{tabularx}
\end{sidewaystable}

\subsubsection{Diverse Strategies for Enhancing Domain Adaptation}
Incorporating ViTs into DA techniques involves multi-modal and varied methods. We have meticulously reviewed these recent advancements, systematically categorizing the diverse strategies with its unique approach employed. In Table \ref{tab:ViT-DA}, we provide a comprehensive overview of the methods utilized in each investigated study, thereby facilitating an easier comparison and understanding for the reader. To offer further insight, for each study, we delve into the various strategies that have been developed. Each strategy showcases a unique approach and possesses distinct strengths, reflecting the diverse nature of the field. Following is an explanation of these methods for a deeper insight:
In the evolving landscape of domain adaptation, \textit{Adversarial Learning (ADL)}, for instance, harnesses adversarial networks to minimize discrepancies between different domains, creating a competitive scenario where models continually improve their domain invariance and adaptability. In contrast, \textit{Cross-DA (CRD)} tackles the challenge of transferring knowledge from a source domain to a target domain, effectively handling the variances in data distributions.

Adding to the diversity, Using \textit{Visual Prompts (VisProp)} leverages visual cues, enriching the learning process, especially in vision-based tasks. This method brings a novel perspective, guiding models through complex visual landscapes. Meanwhile, \textit{Self-Supervised Learning (SSL)} takes a different route by extracting learning signals directly from the data, eliminating the need for labeled datasets and enabling models to uncover underlying patterns in an unsupervised manner.

The fusion of different architectural paradigms, as seen in Hybrid Networks \textit{combining ViTs with CNN (ViT+CNN)}, brings forth the best of both worlds, the perceptual strengths of CNNs and the relational prowess of transformers. \textit{Knowledge Distillation (KD)} enables a smaller, more efficient model to learn from a larger, more complex one, encapsulating the essence of efficient learning.

In scenarios where access to the original source data is restricted, \textit{Source-Free DA (SFDA)} emerges as a crucial strategy. It relies on the model's inherent knowledge and the characteristics of the target domain, showcasing adaptability in constrained environments. Complementing this, \textit{Test-Time Adaptation (TTA)} ensures that models remain flexible and adaptable even during the inference phase, crucial for dealing with evolving data landscapes.

The adaptation techniques can be further nuanced based on class overlap scenarios between the source and target domains, leading to \textit{Closed-Set, Partial-Set, and Open-Set Adaptation (CPO)}. Each addresses a specific kind of overlap, from complete to none, reflecting the diverse challenges in domain adaptation. \textit{Pseudo Label Refinement(PLR)} , on the other hand, enhances the reliability of labels in unsupervised settings, refining the model-generated labels for better accuracy.
Lastly, \textit{Contrastive Learning (CL)}, by distinguishing between similar and dissimilar data points, offers a robust way for models to learn distinctive features, essential for tasks like classification and clustering. For methods like game theory, which are sparsely used, we have categorized them under \textit{Aditional Emerging Methods (AEM)} to provide a comprehensive overview.
\begin{sidewaystable}
\caption{Comprehensive Summary of Techniques in ViTs for DA: ADL (Adversarial Learning), CRD (Cross-DA), VisProp (Visual Prompts), SSL (Self-Supervised Learning/Semi-Supervised Learning), ViT+CNN (Hybrid Networks), KD (Knowledge Distillation), SFDA (Source-Free DA), TTA (Test-Time Adaptation), CPO (Closed-Set, Partial-Set, Open-Set Adaptation), PLR (Pseudo Label Refinement), CL (Contrastive Learning), and Additional Emerging Methods (AEM).} \label{tab:ViT-DA}
\begin{tabularx}{\textheight}{C{2cm} C{1cm} C{1cm} C{1cm} C{1cm} C{1cm} C{1cm} C{1cm} C{1cm} C{1cm} C{1cm} C{1cm} C{1cm}}
\toprule
\textbf{Study} & \textbf{ADL} & \textbf{CRD} & \textbf{VisProp} & \textbf{SSL} & \textbf{ViT-CNN} & \textbf{KD} & \textbf{SFDA} & \textbf{TTA} & \textbf{CPO} & \textbf{PLR} & \textbf{CL} & \textbf{AEM} \\
\midrule
CDTRANS \cite{xu2021cdtrans} & & \checkmark & & & & & & & & \checkmark & & \\
\hline
\\
TransDA \cite{yang2021transformer} & & & & \checkmark & \checkmark & \checkmark & \checkmark & & \checkmark & & & \\
\hline
\\
GE-ViT\cite{zhang2022delving} & \checkmark & & & \checkmark & & & & & & & & \checkmark \\
\hline
\\
TVT\cite{yang2023tvt} & \checkmark & \checkmark & & \checkmark & & & & & & & & \\
\hline
\\
BCAT \cite{wang2022domain} & & \checkmark & & & & \checkmark & & & & & & \\
\hline
\\
SSRT \cite{sun2022safe} & \checkmark & & & \checkmark & & & & & & \checkmark & & \\
\hline
\\
DoT \cite{ma2022making} & & & & & & & & & & \checkmark & \checkmark & \\
\hline
\\
PMTarns \cite{zhu2023patch} & & \checkmark & & \checkmark & & & & & & & & \checkmark \\
\hline
\\
DOT \cite{chuan2022towards} & & \checkmark & & & \checkmark & & & & & & & \\
\hline
\\
DePT \cite{gao2022visual} & & & \checkmark & \checkmark & & & & \checkmark & & \checkmark & & \\
\hline
\\
BeiT \cite{tayyab2021pre} & & & & \checkmark & & & & & & & & \\
\hline
\\
SUDA \cite{zhang2022spectral} & \checkmark & & & \checkmark & & & & & & & & \checkmark \\
\hline
\\
SMTDA \cite{kumar2023conmix} & & & & & & \checkmark & \checkmark & & & \checkmark & & \\
\hline
\\
UniDA \cite{zhu2023universal} & & \checkmark & & \checkmark & & & & \checkmark & & & \checkmark & \checkmark \\
\hline
\\
WinTR \cite{ma2021exploiting} & & \checkmark & & & & & & & & \checkmark & \checkmark & \\
\bottomrule
\end{tabularx}
\end{sidewaystable}

\subsection{Vision Transformers in Domain Generalization} In our comprehensive review of the existing researches, we analyzed how ViTs are adapted for the DG process. Based on our findings, we have identified four distinct approaches that encapsulate the common strategies within the literature. we categorized the approaches into four main categories based on our analysis of the literature. These categories are: \textit{Multi-Domain Learning}, \textit{Meta-Learning Approaches}, \textit{Regularization Techniques}, and 
\textit{Data Augmentation Strategies}. In the subsequent section, we will delve into the specifics of the research within each category.
\\

\subsubsection{Multi-Domain Learning} This method involves training ViTs across different types of data or domains. The main goal is to train these models to recognize features that are common across all these domains. By doing this, the models become better at working in new and varied environments they haven't seen before.

INDIGO \cite{mangla2022indigo} is a novel method for enhancing DG. INDIGO stands out by integrating intrinsic modality from large-scale pre-trained vision-language networks with the visual modality of ViTs. This integration, coupled with the fusion of multimodal and visual branches, significantly improves the model's ability to generalize to new, unseen domains. The effectiveness of INDIGO is demonstrated through substantial improvements in DG benchmarks like DomainNet and Office-Home. We will introduce the famous benchmarks in DA and DG in \ref{sec:4}. 

\subsubsection{Meta-Learning Approaches} Meta-learning is an approach for training ViTs to adapt rapidly to new domains with minimal data. By engaging in a variety of learning tasks, ViTs develop the
ability to apply meta-knowledge across different settings, significantly boosting their
adaptability and performance in unseen environments. We categorize several recent studies under meta-learning approaches, including Domain Prompt Learning (DPL) with the DoPrompt algorithm, hybrid architecture with query-memory decoding, and Common-Specific Visual Prompt Tuning (CSVPT), all of which illustrate the effectiveness of these techniques in improving domain generalization.
 In the following paragraphs, we will delve into the details of research focusing on meta-learning approaches.

 \cite{zheng2022prompt} introduces the DoPrompt algorithm, a novel approach in the realm of ViTs for domain generalization. It uniquely incorporates Domain Prompt Learning  and Prompt Adapter Learning, embedding domain-specific knowledge into prompts for each source domain. These prompts are then integrated through a prompt adapter for effective target domain prediction.
 
In \cite{kang2021dynamically}, the authors present an innovative approach for domain generalization using ViTs. It leverages a hybrid architecture that combines domain-specific local experts with transformer-based query-memory decoding. This unique methodology allows for dynamic decoding of source domain knowledge during inference, demonstrating enhanced performance and generalization capabilities on various benchmarks, outperforming existing state-of-the-art methods.
 
 Researchers in \cite{li2022learning}, propose Common-Specific Visual Prompt Tuning, a new method integrating domain-common prompts to capture task context and sample-specific prompts to address data distribution variations, enabled by a trainable prompt-generating module (PGM). This approach is specifically tailored for effective adaptation to unknown testing domains, significantly enhancing out-of-distribution generalization in image classification tasks.

\subsubsection{Regularization Techniques} Regularization methods are essential for preventing overfitting and promoting the learning of generalized features. These methods impose various constraints during training to ensure that the learned features are not overly specific to the source domain, thus improving the model's performance on unseen target domains. Regularization techniques such as self-distillation, cross-attention mechanisms, and test-time adjustments have been shown to significantly enhance the generalization capabilities of ViTs. By encouraging the model to learn broadly applicable features, these methods help to mitigate the impact of domain shifts.

Researchers introduce a Self-Distillation for ViTs (SDViT) approach \cite{sultana2022self}, aiming to mitigate overfitting to source domains. This technique utilizes non-zero entropy supervisory signals in intermediate transformer blocks, encouraging the model to learn features that are broadly applicable and generalizable. The modular and plug-and-play nature of this approach seamlessly integrates into ViTs without adding new parameters or significant training overhead. This research is aptly classified under Regularization Techniques in the taxonomy of DG using ViTs, as the self-distillation strategy aligns with the goal of preventing overfitting and promoting domain-agnostic generalization.

This study \cite{liu2022empirical} proposes a Cross Attention for DG (CADG) model. The model uses cross attention to tackle the distribution shifts problem inherent in DG, extracting stable representations for classification across multiple domains. Its focus on using cross-attention to align features from different distributions, a strategy that enhances stability and generalization capabilities across domains, puts it under the regularization part.

Researchers in \cite{singh2023robust} center on boosting DG through Intermediate-Block and Augmentation-Guided Self-Distillation. The proposed method incorporates self-distillation techniques to boost the robustness and generalization of ViTs, particularly focusing on improving performance in unseen domains. This approach has shown promising results on various benchmark datasets, and it has a commitment to leveraging self-distillation to prevent over-fitting and foster generalization across varied domains.

Test-Time Adjustment (T3A) \cite{iwasawa2021test} proposes an optimization-free method for adjusting the classifier at test time using pseudo-prototype representations derived from online unlabeled data. This approach aims to robustify the model to unknown distribution shifts.

\subsubsection{Data Augmentation Strategies} Data augmentation strategies are applied to increase the diversity and robustness of training datasets. By artificially expanding the training data, these methods help ViTs to learn more generalized and adaptable features, improving their performance on unseen target domains. Advanced data augmentation techniques, including synthetic data generation, spatial transformation, and token-level feature stylization, have shown significant promise in enhancing the generalization capabilities of ViTs. By introducing variability in the training data, these methods help to mitigate the impact of domain shifts and improve model robustness.

The researchers introduce a novel concept known as Token-Level Feature Stylization (TFS-ViT) \cite{noori2023tfs}. This method transforms token features by blending normalization statistics from various domains and applies attention-aware stylization based on the attention maps of class tokens. Aimed at improving ViTs' ability to adapt to domain shifts and handle unseen data, this approach is a prime example of data augmentation strategies in DG using ViTs. TFS-ViT's emphasis on feature transformation and utilizing diverse domain data is an advanced data augmentation technique, aimed at enriching training data variety for enhanced DG.

\cite{kang2021discovering} explores a unique approach to DG by focusing on spatial relationships within image features. The proposed hybrid architecture (ConvTran) merges CNNs and ViTs, targeting both local and global feature dynamics. The methodology is aimed at learning global feature structures through the spatial interplay of local elements, with the aim of generalization. In terms of its relation to data augmentation, the idea of leveraging spatial interplay for categorization in data augmentation is rooted in a comprehension of how image features interact spatially allows a model to better adapt to and perform in novel, previously unseen domains. ConvTran enhances the model’s ability to process and generalize across different domains by learning and incorporating global spatial relationships, aligning with strategies aimed at augmenting training data diversity for better generalization.

The research \cite{schwonberg2023augmentation} carefully examines multiple image augmentation techniques to determine their effectiveness in promoting DG, specifically within the context of semantic segmentation. This investigation includes experiments utilizing the DAFormer architecture  \cite{hoyer2022daformer}, showcasing the wide-ranging applicability of these augmentations across various models. It highlights how important it is to carefully check different ways of changing images. It emphasizes the importance of evaluating a variety of image augmentation strategies, considering that carefully selected data augmentations are essential for improving the generalization abilities of models.

In conclusion, table \ref{tab:Design_highlights_ViT_DG} presents representative works from recent research that have employed ViTs for DG. These selected studies highlight the adaptability and potential of ViTs in improving models' ability to generalize across various domains.

\small 
\begin{longtable}{ L{2cm}  L{1.5cm} L{4cm} L{2.5cm} L{1.5cm} }
\caption{Representative Works of ViTs for DG}
\label{tab:Design_highlights_ViT_DG}\\
    \hline
    \makecell{Category} & \makecell{Method} & \makecell{Design Highlights} & \makecell{Training \\ Strategies} & \makecell{Publication} \\
    \hline
    \endfirsthead
    \hline
    \centering
    \makecell{Category} & \makecell{Method} & \makecell{Design Highlights} & \makecell{Training \\ Strategies} & \makecell{Publication} \\
    \hline
    \endhead
  Meta-Learning Approaches & CSVPT\cite{li2022learning} & Boosts OOD generalization with dynamically generated domain-invariant and variant prompts via a trainable module, improving adaptability across datasets & Cross Entropy & ACCV 2022 \\
  \hline
   Regularization Techniques & SDViT\cite{sultana2022self} & Reduces overfitting by using self-distillation, entropy-based signals, and a modular approach, aiming for better learning across different domains & Cross Entropy, KL Divergence & ACCV 2022 \\
   \\
    & T3A\cite{iwasawa2021test} & Enhances DG by updating linear classifiers with online-generated pseudo-prototypes, offering robustness in varying environments without back-propagation & Cross Entropy, Pseudo Label Refinement & NeurIPS 2021\\
    \hline
\end{longtable}
\normalsize 

\subsubsection{Diverse Strategies for Enhancing Domain Generalization} Building on our discussion from the DA section, we now shift our focus to exploring the various strategies employed in DG. Similar to DA, DG also encompasses a spectrum of methods, adapting known DG methods for ViTs, each tailored to enhance the model's capability to generalize across unseen domains. Here, we delve into these diverse techniques, outlining their unique features and roles in the context of DG. Table \ref{tab:ViT-DG} summarizes the strategies employed in research to address DG challenge through the integration of ViT architecture.

\textit{Domain Synthesis (DST)} creates artificial training domains to enhance the model's generalization capability across unseen environments. \textit{Self Distillation (SD)} leverages the model's own outputs to refine and improve its learning process. \textit{Class Guided Feature Learning (CGFL)} focuses on extracting features based on class-specific information to improve classification accuracy. \textit{Adaptive Learning (ADPL)} dynamically adjusts the learning strategy based on the specifics of each domain. \textit{ViT-CNN} Hybrid Networks combine the strengths of ViTs and CNN for robust feature extraction. \textit{Feature Augmentation/Feature Learning (FAug)} enhances feature sets to improve model robustness against varying inputs. \textit{Prompt-Learning (PL)} employs guiding prompts to direct the learning process, particularly useful in language and vision tasks. \textit{Cross Domain (CRD)} learning involves training models across diverse domains to improve adaptability. \textit{Source Domain Knowledge Decoding (SDKD)} decodes and transfers knowledge from the source domain to enhance generalization. \textit{Knowledge Distillation (KD)} transfers knowledge from a larger, complex model to a smaller, more efficient one. \textit{Source-Free DA (SFDA)} adapts models to new domains without relying on source domain data, crucial for privacy-sensitive applications. \textit{Multi Modal Learning (MML)} uses multiple types of data inputs, such as visual and textual, to improve learning comprehensiveness. \textit{Test-Time Adaptation (TTA)} adjusts the model during inference to adapt to new environments, ensuring robust performance on unseen data.

To conclude this chapter, we have discussed various tables analyzing different ViT-based methods for DA and DG from multiple viewpoints. For a summary of the advantages and limitations of these methods, refer to Table \ref{tab:Domain_Adaptation_Generalization_ViT}
. The tables in this chapter offer a comprehensive overview and detailed analysis of the studies from different perspectives.
\begin{sidewaystable}
\caption{Comprehensive Summary of Techniques in ViTs for DG: DST (Domain Synthesis), SD (Self Distillation), CGFL (Class Guided Feature Learning), ADPL (Adaptive Learning), ViT-CNN (Hybrid Networks), FAug (Feature Augmentation/Feature Learning), PL (Prompt-Learning), CRD (Cross Domain), ViT+CNN (Hybrid Networks), SDKD (Source Domain Knowledge Decoding), KD (Knowledge Distillation), SFDA (Source-Free DA), MML (Multi Modal Learning), and TTA (Test-Time Adaptation).} \label{tab:ViT-DG}
\begin{tabularx}{\textheight}{C{2cm} C{1cm} C{1cm} C{1cm}C{1cm} C{1cm} C{1cm} C{1.5cm} C{1cm} C{1cm} C{1cm}C{1cm}}
\toprule
\textbf{Study} & \textbf{DST} & \textbf{SD} & \textbf{CGFL} & \textbf{ADPL} & \textbf{ViT-CNN} & \textbf{FAug/FL} & \textbf{PL} & \textbf{CRD} & \textbf{SFKD} & \textbf{MML} & \textbf{TTA} \\
\midrule
SDViT \cite{sultana2022self} &  & \checkmark & \checkmark &  &  &   &   &  &  &   & \\
\hline
\\
TFS-ViT \cite{noori2023tfs} & \checkmark & & \checkmark & & & \checkmark &   & &  & & \\
\hline
\\
DoPrompt \cite{zheng2022prompt} & & & & \checkmark & & & \checkmark  & &  & & \\
\hline
\\
ConvTran \cite{kang2021discovering} & & & & & \checkmark & & & &  & & \\
\hline
\\
D$^2$SDK \cite{kang2021dynamically} & & & & & \checkmark & & & \checkmark & \checkmark & & \\
\hline
\\
INDIGO \cite{mangla2022indigo} & & & & & & & &  & & \checkmark & \checkmark \\
\hline
\\
CSVPT \cite{li2022learning} & & & & \checkmark & & & \checkmark &  & & & \\
\hline
\\
CADG \cite{dai2022cadg} & & & \checkmark & & & \checkmark &  & \checkmark  & & & \\
\hline
\\
RRLD \cite{singh2023robust} & & \checkmark & & & & \checkmark &  &   & & & \checkmark \\
\hline
\\
T3A+ViT \cite{iwasawa2021test} & & & & & & \checkmark &  &   & & \checkmark & \\
\bottomrule
\end{tabularx}
\end{sidewaystable}

\begin{sidewaystable}
\caption{A summary of key features, advantages and challenges of different Transformers based methods in domain adaptation and domain generalization approaches.}\label{tab:Domain_Adaptation_Generalization_ViT}
\begin{tabularx}{\textheight}{L{2cm} L{2cm} Y L{4cm}  L{3cm} L{3cm}}
\toprule
\makecell{Approaches} & \makecell{Category} & \makecell{Key Features} & \makecell{Studies} & \makecell{Advantages} & \makecell{Challenges} \\
\midrule
\multirow{4}{*}{\makecell{Domain\\ Adaptation}} & Feature-Level adaptation& Aligning feature distributions between domains & DOT \cite{ma2022making}, TRANS-DA \cite{ye2023learning}, DoT \cite{chuan2022towards}, SUDA \cite{zhang2022spectral}, SAMB \cite{li2022semantic}, DePT \cite{gao2022visual}, CTTA \cite{gan2023decorate} & Reduces domain discrepancy & May require complex feature engineering \\
\\
& Instance-Level adaptation & Selecting/weighting data points relevant to target & SF-OSDA \cite{vray2023source} & Enhances relevant feature learning & Computationally intensive \\
\\
& Model-Level adaptation &  Modifying model architecture for better adaptability &  GE-ViT \cite{zhang2022delving}, TransDA\cite{yang2021transformer}, BeiT \cite{tayyab2021pre}, TFC \cite{wang2022tfc} & Directly addresses model limitations & Necessitates redesigning of model architecture\\
\\
& Hybrid Approaches  & Integrating feature, instance, and model adaptation techniques &  CDTRANS \cite{xu2021cdtrans}, TVT \cite{yang2023tvt}, SSRT \cite{sun2022safe}, BCAT \cite{wang2022domain}, PMTrans \cite{zhu2023patch}, UniAM \cite{you2019universal}, CoNMix \cite{kumar2023conmix}, WinTR \cite{ma2021exploiting}& Leverages strengths of multiple approaches & Complexity in integrating different methods \\
\midrule
\multirow{4}{*}{\makecell{Domain\\ Generalization}} & Multi-Domain Learning & Training on multiple source Domains and capturing features invariant across multiple domains & INDIGO \cite{mangla2022indigo} & Improves feature robustness by leveraging diverse data & Requires diverse source domains \\
\\
& Meta-Learning &  Meta-training and meta-testing to adapt to new domains with minimal data & DPL \cite{zheng2022prompt}, MoE  \cite{kang2021dynamically}, CSVPT \cite{li2022learning} & Enhances adaptation speed and effectiveness in new environments & May require extensive meta-training \\
\\
& Regularization & Imposing constraints during training and preventing overfit to source domains & SDViT \cite{sultana2022self}, CADG \cite{liu2022empirical}, RRLD \cite{singh2023robust}, T3A \cite{iwasawa2021test} & Boosts model robustness and ensures stable performance across domains & Balancing regularization strength, implementation complexity, performance trade-offs \\
\\
& Data Augmentation & Generates synthetic variations to simulate target domain characteristics & TFS-ViT \cite{noori2023tfs}, ConvTran \cite{kang2021discovering}, AugDA\cite{schwonberg2023augmentation}, DAFormer\cite{hoyer2022daformer} & Provides diverse training scenarios, improving model adaptability & Ensuring augmented data relevance \\
\bottomrule
\end{tabularx}
\end{sidewaystable}

\section{Applications Beyond the Image Recognition}
\label{sec:4}
Most of the research discussed in section \ref{sec:3} primarily focuses on image recognition tasks. However, these methods have the potential for broader application across various domains. A substantial portion of the studies explores applications extending beyond image recognition to other fields. We have divided these studies into four distinct categories: \textit{semantic segmentation}, which examines the partitioning of images into segments; \textit{action recognition}, focusing on identifying and classifying actions within videos; \textit{face analysis}, which involves detecting and interpreting facial features and expressions; and \textit{medical imaging}, where methods are employed to analyze interpret medical images. In the upcoming sections, we will first briefly discuss benchmarking datasets commonly used in the research, providing a foundation for understanding their methodologies. 
\\

\begin{figure*}
\centering
\includegraphics[width=1\linewidth]{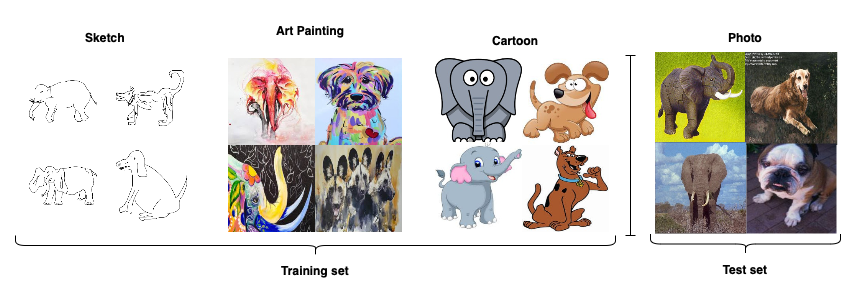}
\caption{ Examples from the PACS dataset \cite{li2017deeper} demonstrate distribution shift scenarios. The training set includes images from sketch, cartoon, and art painting domains, while the testing dataset consists of real images, highlighting the challenges of distribution shift in PACS dataset.}
\label{dg-image}
\end{figure*}

\textbf{Benchmarking Datasets:}\\
In DA and DG approaches, a key focus is on how models perform on datasets with distribution shifts. Such benchmarks are crucial in determining the robustness and adaptability of models, against real-world data variation. The methods for DA/DG, are tested across diverse datasets like VLCS \cite{fang2013unbiased}, Office-31 \cite{saenko2010adapting}, PACS \cite{li2017deeper}, OfficeHome \cite{venkateswara2017deep}, DomainNet \cite{peng2019moment}, and ImageNet-Sketch \cite{wang2019learning}. These evaluations also include scenarios like synthetic-vs-real \cite{fang2013unbiased}, artificial corruptions \cite{hendrycks2019benchmarking}, and diverse data sources \cite{rebuffi2017learning}. To illustrate the distribution shift, samples from the PACS dataset are depicted in Figure \ref{dg-image}.

\subsection{Semantic Segmentation}
In the field of semantic segmentation, a crucial challenge is the limited generalization of DNNs to unseen domains, exacerbated by the high costs and effort required for manual annotation in new domains. This challenge highlights the need for developing new methods and modern visual models to adapt to new domains without extensive labeling, addressing distribution shifts effectively. The shift from synthetic to real data is particularly critical, as it allows for leveraging simulation environments. In this section we offer an in-depth overview of the latest progress in this research area of ViTs, focusing on the sustained efforts and the key unresolved issues that hinder the broader application of ViTs in semantic segmentation across various domains.

DAFormer \cite{hoyer2022daformer} stands out as a foundational work in using ViTs for UDA, presenting groundbreaking contributions at both the method and architecture levels. This approach significantly improved the state-of-the-art (SOTA) performance, surpassing ProDA \cite{zhang2021prototypical}, by more than 10\% mIoU. The architecture of DAFormer is based on SegFormer \cite{xie2021segformer}, which is utilized as the encoder architecture. It incorporates two established methods from segmentation DNNs. DAFormer first introduces skip connections between the encoder and decoder for improved transfer of low-level knowledge. It then employs an ASPP-like \cite{chen2018encoder} fusion, processing stacked encoder outputs at various levels with different dilation rates, aiming to increase the receptive field. At the method level, DAFormer adapts known UDA methods for CNNs, including self-training with a teacher-student framework, strong augmentations, and softmax-based confidence weighting. Additional features include rare class sampling in the source domain and a feature distance loss to pre-trained ImageNet features. An interesting observation made in the study is the potential benefit of learning rate warm-up methods for UDA.

Building directly on the contributions of DAFormer \cite{hoyer2022daformer}, HRDA \cite{hoyer2022hrda} marks a substantial progress in the application of ViT models. Its primary contribution is a scale attention mechanism that processes high and low-resolution inputs, allocating attention scores to prioritize one over the other based on class and object scales. This method facilitates better extraction of contextual information from smaller sections of images and includes self-training using a sliding window for pseudo-label generation. While HRDA further enhances DAFormer's performance, there remains a gap to be bridged.

TransDA \cite{chen2022smoothing} addresses a high-frequency problem identified in ViTs, using the Swin transformer \cite{liu2021swin} architecture. It shows that target pseudo labels and features change more frequently and significantly over iterations compared to a ResNet-101, suggesting this issue is specific to ViT networks. TransDA's solution includes feature and pseudo label smoothing using a momentum network, combined with self-training and weighted adversarial output adaptation, similar to CNN's teacher-student approaches.
Zhang et al. in \cite{zhang2022behind} introduce Trans4PASS+, an advanced model tackling the challenges of panoramic semantic segmentation. This model addresses image distortions and object deformations typical in panoramic images, utilizing Deformable Patch Embedding (DPE) and Deformable MLP (DMLPv2) modules. Additionally, it features a Mutual Prototypical Adaptation (MPA) strategy for UDA in panoramic segmentation, enhancing performance in both indoor and outdoor scenarios. The paper also contributes a new dataset, SynPASS, to facilitate Synthetic-to-Real (SYN2REAL) adaptation in panoramic imagery.

In the context of these developments, Ding et al. introduce HGFormer \cite{ding2023hgformer}, a novel approach for domain generalization in semantic segmentation. HGFormer groups pixels into part-level masks before assembling them into whole-level masks. This hierarchical strategy significantly enhances the robustness of segmentation against domain shifts by combining detailed and broader image features. HGFormer's effectiveness is demonstrated through various cross-domain experimental setups, showcasing its superiority over traditional per-pixel classification and flat-grouping transformers.

Alongside these innovative approaches, a growing number of studies, such as ProCST \cite{ettedgui2022procst}, are evaluating ViT networks. ProCST applies hybrid adaptation with style transfer in the input space, in conjunction with DAFormer \cite{hoyer2022daformer}and HRDA\cite{hoyer2022hrda}. Recently \cite{schwonberg2023augmentation} delves into the efficacy of simple image-style randomization and augmentation techniques, such as blur, noise, and color jitter, for enhancing the generalization of DNNs in semantic segmentation tasks. The study is pivotal in its systematic evaluation of these augmentations, demonstrating that even basic modifications can significantly improve network performance on unseen domains. Notably, the paper reveals that combinations of multiple augmentations rival the complexity and effectiveness of state-of-the-art domain generalization methods. Employing architectures like ResNet-101 and the ViT DAFormer, the research achieves remarkable results, with performance on the synthetic-to-real domain shift between Synthia and Cityscapes datasets reaching up to 44.2\% mIoU. Rizzoli et al. introduce MISFIT \cite{rizzoli2023source}, a novel framework for multimodal source-free domain adaptation in semantic segmentation. This method innovatively fuses RGB and depth data at multiple stages in a ViT architecture. Key features include input-level depth stylization for domain alignment, cross-modality attention for mixed feature extraction, and a depth-based entropy minimization strategy for adaptively weighting regions at different distances. MISFIT, as the first RGB-D ViT approach for source-free semantic segmentation, demonstrates notable improvements in robustness and adaptability across varied domains. Various other works integrate their methods with the DAFormer framework, incrementally improving performance \cite{zhou2022context, xie2023sepico, vayyat2022cluda, du2022learning}, though not surpassing HRDA. Notably, CLUDA \cite{vayyat2022cluda} builds upon HRDA, further improving its performance.

\subsection{Action Recognition}
In the field of surveillance video analysis, a growing area of interest is domain-adapted action recognition. This involves training action recognition systems in one environment (the source domain) and applying them in another with distinct viewpoints and characteristics (the target domain). This emerging research topic addresses the challenges posed by these environmental differences \cite{gao2021novel}. In the context of domain adaptation for action recognition, while source datasets provide action labels, these labels are not available for the target dataset. Consequently, evaluating the performance on the target dataset poses a challenge due to the absence of these labels \cite{tang2022learning}. In the field of RGB-based action recognition tasks, transformer-based domain adaptation methods have demonstrated outstanding. UDAVT \cite{da2022unsupervised}, a novel approach in UDA for video action recognition, demonstrates a significant advancement in handling domain shifts in video data. Central to its design is the innovative use of a spatio-temporal transformer architecture, which efficiently captures both spatial and temporal dynamics. The framework is distinguished by its unique alignment loss term, derived from the information bottleneck principle, fostering the learning of domain-invariant features. UDAVT employs a two-phase training process, initially fine-tuning the whole transformer with source data, followed by the adaptation of the temporal transformer using the information bottleneck loss, effectively aligning domain distributions. This approach has shown SOTA performance on challenging UDA benchmarks like HMDB withUCF and Kinetics with NEC-Drone, outperforming existing methods and underscoring the potential of transformers in video analysis. The integration of a queue for recent feature representations further enhances the method's effectiveness, making UDAVT a significant contribution to the field of action recognition in videos.

Lin et al. \cite{Lin_2023_CVPR} introduce ViTTA, a method enhancing action recognition models during test time without retraining. This approach focuses on feature distribution alignment, dynamically adjusting to match test set statistics with those of the training set. A key aspect is its applicability to both convolutional and transformer-based networks. ViTTA also enforces consistency in predictions over temporally augmented video views, a strategy that significantly improves performance in scenarios with distribution shifts, showcasing its effectiveness over previous test-time adaptation techniques. Q. Yan and Y. Hu's research \cite{yan2023transformer} introduces a UDA method tailored for skeleton behavior recognition, addressing the challenge of aligning source and target datasets in domain adaptation. Their method employs a spatial-temporal transformer framework with three flows—source, target, and source-target facilitating effective domain alignment and handling variations in joint numbers and positions across datasets. Key to this approach is the use of subsequence encoding and an attention mechanism that emphasizes local joint relationships, thereby enhancing the representation of skeleton behavior. Comprehensive testing on various skeleton datasets shows the superiority of their Spatial-Temporal Transformer-based DA (STT-DA) method, underscoring its effectiveness in managing the complexities of domain adaptation in skeleton behavior recognition. The concept of applying transformers for skeleton-based action recognition in DA is viewed as a promising and potentially impactful direction in this field of study \cite{xin2023transformer}.

\subsection{Face Analysis}
 Face anti-spoofing (FAS) is a crucial aspect of biometric security systems, addressing the challenge of distinguishing between genuine and fake facial representations \cite{zou2023adversarial, sarker2024enhanced}. 
Recently WACV 2023 research \cite{liao2023domain}introduced a new approach for FAS using ViTs. The authors proposed the Domain-invariant ViT (DiVT) which employs two specific losses to enhance generalizability. The losses include a concentration loss for learning domain-invariant representations by aggregating features of real face data, and a separation loss to differentiate each type of attack across domains. The study highlights the effectiveness of transformers in capturing long-range dependencies and globally distributed cues, crucial for FAS tasks. It also addresses the large model size and computational resource issues commonly associated with transformer models by adopting a lightweight transformer model, MobileViT. The proposed approach differs from previous methods by focusing on the core characteristics of real faces' features and unifying attack types across domains, leading to improved performance and efficiency in FAS applications. In addition, researchers in \cite{zou2023adversarial} explored FAS in surveillance contexts, where image quality varies widely. The paper introduces an Adversarial DG Network (ADGN), which classifies training data into sub-source domains based on image quality scores. It then employs adversarial learning for feature extraction and domain discrimination, achieving quality-invariant features. The approach also integrates transfer learning to mitigate limited training data issues. This innovative method proved effective in surveillance FAS, as evidenced by its performance in the 4th Face Anti-Spoofing Challenge at CVPR 2023.

In addition to the previously discussed work in the domain of FAS, another significant contribution comes from a study focusing on adaptive transformers for robust few-shot cross-domain FAS \cite{huang2022adaptive}. The study presents a novel approach by integrating ensemble adapter modules and feature-wise transformation layers into ViTs, enhancing their adaptability across different domains with minimal examples. This methodology is especially pertinent in scenarios where FAS systems encounter diverse and previously unseen environments. The research demonstrates that this adaptive approach results in both robust and competitive performance in cross-domain FAS, outperforming state-of-the-art methods on several benchmark datasets, even when only a few samples are available from the target domain. This highlights the potential of adaptive transformers in improving the generalizability and effectiveness of FAS systems in real-world applications. In the context of FAS under continual learning, a rehearsal-free method called Domain Continual Learning (DCL) was proposed. It addressed catastrophic forgetting and unseen domain generalization using the Dynamic Central Difference Convolutional Adapter (DCDCA) for ViT models. The Proxy Prototype Contrastive Regularization (PPCR) was utilized to retain previous domain knowledge without using their data, resulting in improved generalization and reduced catastrophic forgetting \cite{cai2023rehearsal}.
\subsection{Medical Imaging}
In the evolving landscape of medical image classification and analysis, ViTs have emerged as a pivotal technology. Their application is primarily aimed at overcoming the challenges of domain generalization, thereby boosting the adaptability of deep learning methods in ever-changing clinical settings and in the face of unseen environments.

Focusing on the critical area of breast cancer detection, where computer-aided systems have shown considerable promise, the use of deep learning has been relatively hampered by a lack of domain generalization. A noteworthy study in this regard explored this issue within the context of mass detection in digital mammography. This research, encompassing a multi-center setup, delved into the analysis of domain shifts and evaluated eight leading detection methods, including those based on transformer models. The findings were significant, revealing that the proposed workflow not only reduced domain shift but also surpassed existing transfer learning techniques in efficacy \cite{garrucho2022domain}.

In the realm of skin lesion recognition \cite{fayyad2023empirical}, where deep learning has made remarkable strides, the issue of overdependence on disease-irrelevant image artifacts raised concerns about generalization. A groundbreaking study introduced EPVT, an innovative domain generalization method, leveraging prompts in ViTs to amalgamate knowledge from various domains. This approach significantly enhanced the models' generalization capabilities across diverse environments \cite{yan2023epvt}.

Another challenging area is medical image segmentation, which struggles due to the inherent variability of medical images \cite{yuan2023effective}. To tackle the limited availability of training datasets, data-efficient ViTs were proposed. However, indiscriminate dataset combinations can result in Negative Knowledge Transfer (NKT). Addressing this, the introduction of MDViT, a multi-domain ViT with domain adapters, marked a significant advancement. This approach allowed for the effective utilization of varied data resources while mitigating NKT, showcasing superior segmentation performance even with an increase in domain diversity \cite{du2023mdvit}.
The robustness against adversarial attacks is a non-negotiable aspect of deep medical diagnosis systems. A novel CNN-Transformer hybrid model was introduced to bolster this robustness and enhance generalization. This model augmented image shape information in high-level feature spaces, smoothing decision boundaries and thereby improving performance on standardized datasets like MedMNIST-2D \cite{manzari2023medvit}.
Liu et al.'s introduction of the Convolutional Swin-Unet (CS-Unet) represents a notable advance in medical image semantic segmentation. By integrating the Convolutional Swin Transformer (CST) block into transformers, they effectively combined multi-head self-attention with convolutions, providing localized spatial context and inductive biases essential for delineating organ boundaries. This model's efficiency and effectiveness, particularly in its capability to surpass existing transformer-based methods without extensive pre-training, underscore the importance of localized spatial modeling in medical imaging \cite{liu2023optimizing}.
A significant stride in domain generalization was achieved with the introduction of BigAug, a deep stacked transformation approach. This method applies extensive data augmentation, simulating domain shifts across MRI and ultrasound imaging. BigAug's application of nine transformations to each image, validated across diverse segmentation tasks and challenge datasets, set a new standard, outperforming traditional augmentation and domain adaptation techniques in unseen domains \cite{zhang2020generalizing}.

The paper by Ayoub et al. \cite{ayoub2023hvit} brings forth innovative techniques in medical image segmentation, addressing the challenge of model generalization across varied clinical environments. Their methodologies significantly enhance the robustness and applicability of deep learning models in medical imaging, ensuring their effectiveness in diverse clinical scenarios.
Furthermore, Li et al. introduced DTNet, the UDA method comprising a dispensed residual transformer block, a multi-scale consistency regularization, and a feature ranking discriminator. This network significantly improved segmentation performance in retinal and cardiac segmentation across different sites and modalities, setting a new benchmark for UDA methods in these applications \cite{li2021dispensed}.
Finally, the use of self-supervised learning with UNETR, incorporating ViTs into a 3D UNET architecture, addressed inaccuracies caused by out-of-distribution medical data. This model's voxel-wise prediction capability enhances the precision of sample-wise predictions \cite{park2021self}.

In reviewing the recent advancements in the application of ViTs for DA and DG in medical imaging, it becomes evident that ViTs are not only versatile but also increasingly effective in addressing domain-specific challenges. The studies surveyed indicate a significant shift towards models that are more adaptable to varying clinical environments, a crucial aspect of real-world medical applications. However, there is an observable need for further refinement in these models to ensure even greater accuracy and robustness, particularly in scenarios involving scarce or highly variable data.
The success of models like EPVT and MDViT in enhancing generalization capabilities across diverse environments suggests a promising direction toward more domain-agnostic approaches. Nevertheless, the balance between model complexity and interpretability remains a key area for future exploration. As the field moves forward, there's potential for integrating more advanced self-supervised learning techniques and exploring hybrid models that combine the strengths of both CNNs and ViTs. This could lead to a new generation of medical imaging tools that are not only more efficient in handling domain shifts but also more accessible and reliable for clinicians in varied healthcare settings.

\begin{figure*}
\centering
\includegraphics[width=1\linewidth]{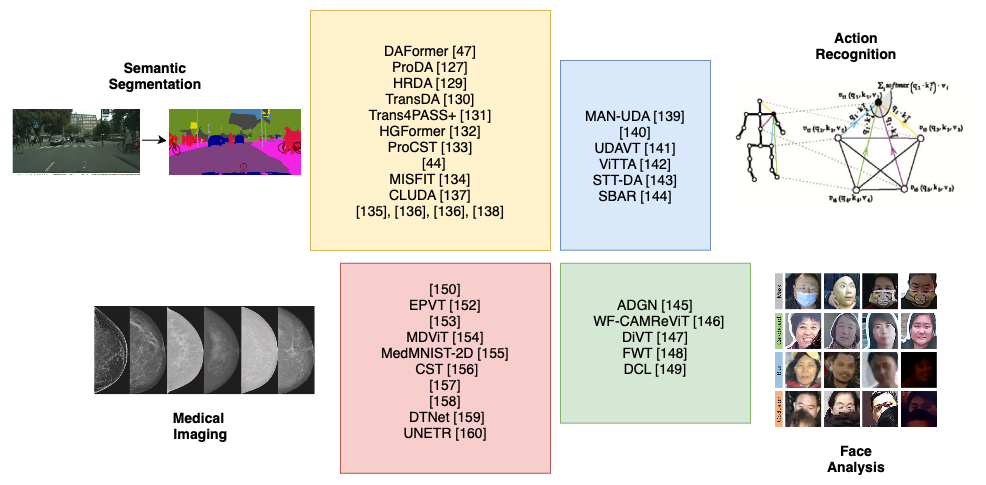}
\caption{Comprehensive categorization of studies leveraging vision transformers for domain adaptation and generalization across diverse tasks, including semantic segmentation \cite{rizzoli2023source}, action recognition \cite{xin2023transformer}, face analysis \cite{zou2023adversarial}, and medical imaging \cite{garrucho2022domain}, extending beyond the image recognition tasks.}
\label{ViT-application}
\end{figure*}

To conclude, this chapter has showcased the extensive utility of ViTs beyond image recognition tasks, highlighting their significant impact in areas such as semantic segmentation, action recognition, face analysis, and medical imaging. Figure \ref{ViT-application} illustrates the categorization of studies utilizing Vision Transformers for domain adaptation and domain generalization across various tasks beyond image recognition. Beyond these fields, ViTs have proven to be highly adaptable and effective in sectors like precision agriculture and autonomous driving \cite{dos2022unsupervised, hasan2022pedestrian}. These researches highlight that the potential of ViTs extends far beyond the initially discussed applications \cite{davuluri2023security}. Their adaptability to different environments and challenges showcases a growing research interest in diverse fields. Future research could explore even more innovative applications, leveraging ViTs' unique ability to handle complex visual tasks and distribution shifts across different industries.

\section{Conclusion, Discussion and Future Research Directions}
\label{sec:5}
This comprehensive review looks at how modern vision networks, ViTs, are used in DA and DG methods to handle distribution shifts. Reviewing the research, we've observed that the transformer’s self-attention mechanisms play a pivotal role in generalizing and adapting to new, distribution-shifted samples, as evidenced by reviewing the experiment results of research in the known shifted datasets including ImageNet-A, ImageNet-C, and Stylized ImageNet. With respect to distribution shifts' handling methods, for DA methods, we've organized the research into four categories: feature-level, instance-level, model-level adaptations, and hybrid approaches. Additionally, we've introduced a new category for studies that combine different strategies with ViTs to tackle distribution shifts. These hybrid networks show the use of various strategies alongside ViTs, leading us to categorize the papers based on these combined approaches. A similar methodological framework was applied to DG, wherein papers are classified based on multi-domain learning, meta-learning, regularization techniques, and data augmentation strategies. Our review is the first to catalog the use of ViTs in DA and DG tasks comprehensively. The tables provided organize and present an overview of the extant literature, demonstrating the burgeoning interest and application of ViTs across various domains. We observed that while the field is rapidly expanding, the literature still shows a sparsity. This has led us to implement additional categorizations for the diverse strategies ViTs employ in DA and DG, enhancing the readability and analytical depth for researchers. These two distinct methods of categorization, by considering various approaches, provide deeper insights and a broader perspective. Finally, we extend our review to applications beyond image recognition, showcasing the versatility of these methods in various domains, and highlighting the potential of ViTs in a broad range of real world applications particularly in critical safety and decision-making scenarios beyond the image recognition tasks.

In our discussion and exploration of various research works, it becomes evident that we are at the nascent stage of developing these modern vision networks. Researchers are increasingly focusing on the characteristics of ViTs across diverse deep learning scenarios. The challenges we faced in compiling and categorizing sparse references have been significant, particularly due to the rapid adoption and development of ViTs. While many papers claim the superiority of ViTs over CNNs, a balanced perspective considering both architectures is essential, depending on their robustness and the features crucial for specific applications. This aspect was evident in our exploration of robustness, where different factors important in generalization and stability were considered, with ViTs sometimes outperforming CNNs in extracting certain features.

While ViTs face challenges stemming from various factors, they are on a trajectory of improvement, much like CNNs were in their early stages. There is a noticeable rise in publications dedicated to modern vision models, emphasizing the advancements being made. As highlighted earlier, the attention mechanism inherent in ViTs significantly enhances their proficiency in handling distribution shifts, marking an advantage over CNNs in this aspect.

In line with the focus of this study, this section outlines potential research areas in the field of tackling distribution shifts. The recommendations take into account the properties of ViTs, as well as DA and DG approaches for managing distribution shifts, and how they can be effectively integrated. ViTs have demonstrated the capacity to surpass previous state-of-the-art methods in certain contexts. However, their superiority in some tasks is not consistently overwhelming, partly due to their reliance on manually designed architectures, which may impede their adaptability, as explored by \cite{lin2022survey}.

Despite the comprehensive nature of this review, it is noteworthy to mention the following limitations with respect to our research. Our review primarily focuses on studies using well-known benchmark datasets, which may limit the generalizability of our findings to other datasets and real-world scenarios not covered in the review. The datasets reviewed may not fully capture the diversity of real-world distribution shifts, potentially overlooking scenarios where ViTs might struggle. Additionally, the review is based on the current state of research, which may evolve with new methodologies and findings. Some promising methods might not have been included due to publication timing. Furthermore, while theoretical and experimental results are extensively reviewed, there is limited practical validation of ViTs in real-world applications within this paper. Certain assumptions were made regarding the performance and applicability of ViTs based on existing literature, which might not hold true in all practical scenarios.

In conclusion, this review not only aggregates and analyzes the role of ViTs within DA and DG frameworks but also outlines potential areas for future research, aimed at creating more robust and versatile deep learning models. As the first survey of its kind, it marks a significant step in understanding and advancing the capabilities of modern vision networks in handling distribution shifts, pointing towards a promising future in this dynamic field. Looking forward, the field faces substantial challenges such as the extensive data requirements and computational intensity of ViTs, alongside a need for real-world, application-specific datasets to validate new DA and DG approaches. To address these limitations and further enhance the applicability of ViTs, future research should focus on several key areas, including expanding the diversity of datasets used, improving practical validation in real-world applications, and keeping pace with evolving methodologies. In the following subsections, we will detail the challenges and outline future research directions accordingly.

\subsection{Data Requirements and Computational Intensity}
Upon investigating the deployment of ViTs within the realms of DA and DG methods, our research has identified a range of challenges that merit closer examination. ViTs have emerged as a pivotal enhancement in computer vision capabilities, promising significant advances across various applications. However, their adoption in real-world scenarios is fraught by considerable challenges, primarily due to their extensive data requirements. The necessity for large-scale datasets, such as JFT-300M, for effective training, underscores this challenge, as \cite{dosovitskiy2020image} have pointed out. This substantial reliance on voluminous, pre-trained models, emphasized by \cite{touvron2021training}, necessitates access to high-quality data, a critical aspect particularly when training ViTs from scratch on more constrained datasets, a scenario highlighted by \cite{liu2021swin}, and recently discussed in \cite{nie2024scopevit}. This dependency presents a significant barrier, wherein the effectiveness of ViTs, upon adaptation through transfer learning and fine-tuning, is closely linked to the quality of pre-existing object detectors. This necessitates a meticulous application of these methodologies, as evidenced by the works of \cite{lu2019vilbert}, \cite{yang2020learning} and scalable approaches like \cite{nie2024scopevit}. 
In addition, a big challenge is when the models need to learn from a very small number of examples \cite{akkaya2024enhancing}, a method known as few-shot learning. Situations that require high safety standards or have very limited data, show how hard it can be to adjust ViTs under different conditions. Recently, new methods like source-free domain adaptation and test time adaptation have shown promise in making models more generalized. Even though researchers made progress in overcoming biases from the source data \cite{akkaya2024enhancing}, they still have a lot to learn about how to manage uncertainty when making these adjustments. This area presents significant opportunities for further research.
Finally, the inherent self-attention mechanisms of ViTs, especially in models with a vast number of trainable parameters, introduce a layer of complexity that demands substantial computational resources. This further complicates their deployment in practical scenarios, as discussed by \cite{lin2021end}.

\subsection{Necessity of New Benchmarks}
With respect to handling distribution shift approaches, they have presented their own set of challenges. The recent VisDA-2021 dataset competition \cite{bashkirova2022visda}, where transformers underpinned the winning solutions, indicates their efficacy in managing robustness against distribution shifts. This observation aligns with findings by \cite{bai2021transformers}, asserting transformers' superior generalization capabilities on target domain samples. While the advancements in performance over conventional baseline backbones, such as CNNs, are commendable, the quest for perfection is ongoing and remains substantially unfulfilled. This gap underscores the necessity for new benchmarks aimed at propelling research on real-world distribution shift approaches further. The limited number of datasets currently employed in DA and DG intensifies the challenge of validating new approaches, highlighting a need for real-world, application-specific datasets. This review, although broad in scope, reveals a prevailing bias towards classification tasks, even when exploring applications beyond image recognition. In the domain of medical imaging, for instance, this bias persists, underscoring the importance of extending the focus of ViTs to encompass a wider array of tasks beyond mere classification tasks.

\subsection{Pre and Post Domain Adaptation Approaches}
In the context of DA, the utilization of pre-domain adaptation (Pre-DA) and post-domain adaptation (Post-DA) strategies plays a pivotal role in enhancing model performance. Pre-DA focuses on preparing models before they are exposed to new domains, aiming to address and bridge domain discrepancies beforehand. Meanwhile, Post-DA strategies are applied after exposure to the new domain \cite{liu2021unsupervised}, with the goal of mitigating accuracy declines. The significance of integrating both Pre-DA and Post-DA approaches has been underscored in recent studies, suggesting that a comprehensive exploration of these strategies could substantially improve the adaptability and effectiveness of models in unfamiliar domains \cite{singhal2023domain}.
Another significant challenge is the lack of effective comparison metrics for certain DA and DG scenarios. The common use of absolute mean Average Precision (mAP) for object detection tasks does not fully capture the subtleties of evaluation metrics, where relative improvements post-DA might be more indicative of success. This highlights a need for robust comparison metrics capable of accommodating the variability inherent in models trained under diverse conditions \cite{bashkirova2022visda}.
\subsection{Uncertainty-Aware Vision Transformers}

In our analysis of ViTs and their proficiency in navigating distribution shifts, we've highlighted their potential to enhance model generalization through various techniques. A particularly promising, yet under explored approach is integrating uncertainty quantification methods with ViTs \cite{guo2024uctnet}. This integration enables models to provide predictions along with confidence levels, making decision-making more transparent. The presence of uncertainties, amplified by distribution shifts, is not merely an additional challenge but a crucial aspect of real-world environments' unpredictability. Employing uncertainty-aware ViTs to detect and improve model generalizability presents a significant research opportunity. Future studies should delve into how uncertainties influence the adaptation and generalization capabilities of ViTs, emphasizing the integration of uncertainty quantification methods. Such investigative efforts are crucial for gaining a thorough understanding of how modern vision networks can exploit uncertainties to enhance the field of domain adaptation and generalization.\\


This review is the first to comprehensively gather recent work on using ViTs to address distribution shifts in DA and DG approaches. The growing number of publications highlights increasing interest and rapid evolution in the field. We see a promising future for ViTs in addressing distribution shifts and aim to guide future research toward creating more robust and versatile deep learning models.
\\
\\

\textbf{Data availability}
The data that support the findings of this study are publicly available and will be provided upon request.

\bibliography{sn-bibliography}

\end{document}